\documentclass[twocolumn]{revtex4-2}
 
\usepackage{graphicx}
\usepackage[all]{xy}
\usepackage{tikzsymbols}
\usepackage{scalerel}
\usepackage{amsthm}
\usepackage{amsmath}
\usepackage{amsfonts}
\usepackage{color}
\usepackage{bbm}
\usepackage{algorithm}
\usepackage{algpseudocode}
\usepackage[colorlinks=true, allcolors=blue]{hyperref}
\usepackage{xr}

\DeclareMathOperator\erf{erf}
\newcommand{\ext}[0]{\mathcal{E}(\sigma)}

\begin{document}
\title{Noise-enabled goal attainment in crowded collectives}

\author{
Lucy Liu,$^{1}$
Justin Werfel,$^{1}$
Federico Toschi,$^{2,3}$
L.~Mahadevan$^{1,4,5}$
}\email{Corresponding author: lmahadev@g.harvard.edu}

\affiliation{$^1$John A. Paulson School of Engineering and Applied Sciences, Harvard University, Cambridge, MA 02138}
\affiliation{$^2$Department of Applied Physics and Science Education, Eindhoven University of Technology, Eindhoven 5600 MB, The Netherlands}
\affiliation{$^3$Istituto per le Applicazioni del Calcolo ``M. Picone'', Consiglio Nazionale delle Ricerche, Rome I-00185, Italy}
\affiliation{$^4$Department of Physics, Harvard University, Cambridge, MA 02138}
\affiliation{$^5$Department of Organismic and Evolutionary Biology, Harvard University, Cambridge, MA 02138}

\begin{abstract}
\noindent \textbf{Significance Statement}: How many cooks in a kitchen make it too crowded to function? Here we tackle the mathematical challenge of predicting how quickly individuals can reach goal locations in a crowded environment. Having agents travel in less direct, more random paths can break up jams and improve the goal attainment rate. We develop estimates for travel times, then use them to find the optimal crowd density and level of randomness in the path that maximize the goals reached.  We  deploy our navigation strategies using robots to determine their efficacy in crowded environments. Our theoretical findings inform the effective design of robot teams and pedestrian spaces, and our experiments demonstrate new possibilities for studying crowd dynamics using robots.

\medskip

\noindent \textbf{Abstract}: In crowded environments, individuals must navigate around other occupants to reach their destinations. Understanding and controlling traffic flows in these spaces is relevant for coordinating robot swarms and designing infrastructure for dense populations. Here, we use simulations, theory, and experiments to study how adding stochasticity to agent motion can reduce traffic jams and help agents travel more quickly to prescribed goals. A computational approach reveals the collective behavior. Above a critical noise level, large jams do not persist. From this observation, we analytically approximate the swarm's goal attainment rate, which allows us to solve for the agent density and noise level that maximize the goals reached. Robotic experiments corroborate the behaviors observed in our simulated and theoretical results. Finally, we compare simple, local navigation approaches with a sophisticated but computationally costly central planner. A simple reactive scheme performs well up to moderate densities and is far more computationally efficient than a planner, motivating further research into robust, decentralized navigation methods for crowded environments. By integrating ideas from physics and engineering using simulations, theory, and experiments, our work identifies new directions for emergent traffic research.
\end{abstract}

\maketitle

Consider a robot team with a time-sensitive distributed task such as assembling a machine, fulfilling orders in a warehouse, or cleaning up hazardous debris. Robots must transport items to specific goal locations. When the space is relatively empty, adding robots is advantageous: several robots together work faster than a lone one. However, adding too many robots will lead to traffic that slows the entire team down. Systems of many interacting agents are challenging to analyze mathematically. Approximation of a global metric, like how quickly a collective reaches goals, has been an elusive result in emergent traffic. Achieving it would enable optimized system design, including the ability to determine the optimal team size for a task.

Emergent traffic patterns like jam formation, laning, and various transitions between ordered and disordered behavior have been studied in diverse settings spanning car traffic~\cite{lighthill1955kinematic,newell1993simplified}, colloids and bacteria \cite{cates2015motility}, robots \cite{marcolino2009traffic, soriano2017avoiding}, ants \cite{poissonnier2019experimental, couzin2003self}, and humans \cite{corbetta2016continuous, bacik2023lane, iyer2024directed}. In these systems, the simple constraint that two agents cannot occupy the same location at the same time, so that agents must stop or slow down in high-traffic regions, produces a set of rich and interesting phenomena. For instance, it is known that collectives of random walkers with exclusion constraints alone and no attraction can self-organize into large jams \cite{cates2015motility}, and that ants or pedestrians following simple rules can self-organize into lanes \cite{bacik2023lane, couzin2003self}.  More recently, interest has grown in studying active particles with minimal intelligence, such as agents with sensing cones \cite{negi2022emergent} and individual goal locations \cite{casiulis2021self}. A natural question is how to maximize the efficiency of target attainment in a crowded environment. How many agents should be used, and what navigation strategy should they follow? How effectively does noisy motion, the simplest mechanism for exiting jams, enable flow? How do complex global planners compare to a simple reflexive rule each agent follows on its own? 

These questions sit at the nexus of decision making, robotics, animal behavior and statistical physics. In robotics, the problem of computing noncolliding paths between multiple agents and their goals has led to the Multi-Agent Pathfinding (MAPF) problem, for which finding time-optimal paths is NP-hard \cite{yu2013structure}. One baseline algorithm for MAPF is Cooperative A* \cite{silver2005cooperative}, a greedy algorithm where robots plan paths one after the other and can access full information about already-planned paths. Cooperative A* does not guarantee optimal solutions, so subsequent improvements have focused on speed and completeness \cite{standley2010finding}. Variations of MAPF may allow robots to swap goals or may extend the problem to a lifelong version where new goals are continually added \cite{ma2017lifelong, li2021lifelong}. In animal behavior, it is now well accepted that simple local rules for carrying out collective tasks are often far more robust than global planning; this is most clearly visible in social insects that work as super-organisms~\cite{couzin2003self, peleg2018collective, heyde2021self}. In physics, recent work on active particles with noisy trajectories and independent random goals \cite{casiulis2021self} has shown that they reach goals most efficiently   when the noise level  is just above ``the edge of jamming."  Motivated by these and similar studies, we take interest in approximating system-level outcomes to help select optimal strategies for target attainment in crowded environments. 

Our setting draws inspiration from decentralized robot collectives where task execution is possible without any central planner~\cite{giomi2013swarming, prasath2022dynamics, rubenstein2014programmable, werfel2014designing}. A decentralized and noisy approach is relevant to agents which are hard to precisely control, like bacteria, insects, humans, microrobots, or robots in low-communication and low-visibility settings. Because the setting is highly stochastic, we are able to derive analytical results using probability theory.

Robot paths to the goal are entirely determined by local sensing and random noise. Each robot starts at a random location and receives a random goal to travel to. The robot always knows its goal's location and direction, and it homes in toward its goal in a random walk with rotational noise. Robots cease forward motion when a neighbor is detected within their forward sensing cones. The noisy walks allow robots to move stochastically around each other, and the swarm can be tuned by adjusting the swarm size and the rotational noise level. Noisy motion helps unjam traffic, but it also makes robots' paths indirect and inefficient. We measure a swarm's success using the goal attainment rate $G$, the average goals reached per second by all team members combined. 
 
To address the question of what team size and noise level will maximize $G$, we begin with a qualitative understanding of how system behavior changes as we vary swarm size $n$ and noise level $\sigma$. We use our observations to derive estimates for the length of a noisy walk, the jam escape time, the critical noise level, and other useful quantities in a theoretical setting. We assemble our derivations into an analytical approximation for $G$ as a function of $n$. Numerical optimization of $G$ yields an estimate for the optimal $n$ and $\sigma$. Next, we implement the noisy navigation method on a swarm of physical robots. Finally, we compare our minimal, noisy navigation method to two more intelligent methods: a slightly more intelligent walk where noise is selectively applied only when robots encounter obstacles, and a centralized planner with full global knowledge.

\section*{Agentic model simulations}
\begin{figure*}
    \centering
        \includegraphics[width=0.7\linewidth]{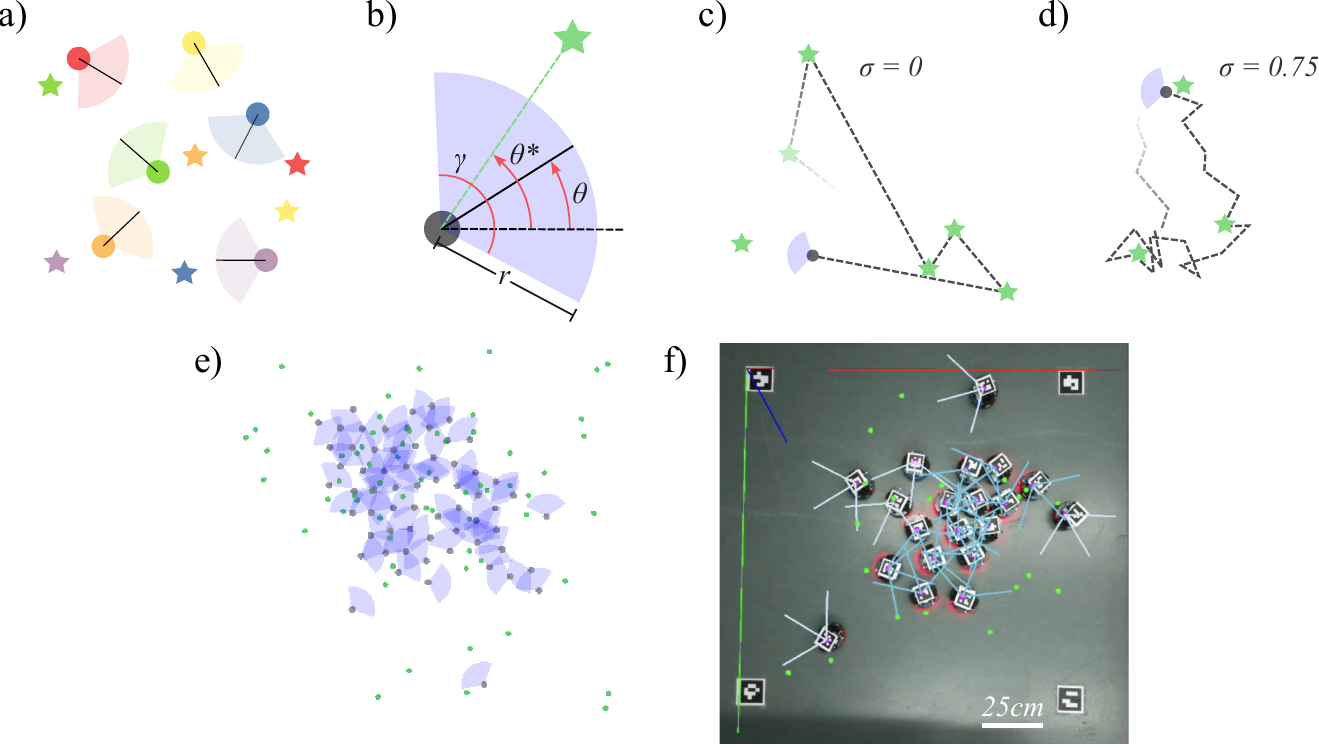}
    \caption{
    \textbf{Noisy swarm setting.}
    \textbf{a)} Robots begin in random positions and individually receive random goal positions (star shapes of matching color), which represent locations for dropping off deliveries or construction parts. See SI Video 1 for animated examples of robot trajectories.
    \textbf{b)} Robots cease forward motion, but can still rotate in place, when an occupant is detected in their sensing cones (blue). The sensing cone is a $\gamma = 120^{\circ}$ sector of a circle with radius $r$. At time $t$, robots have heading $\theta(t)$ and an optimal travel angle $\theta^*(t)$ to the goal (green star). 
    \textbf{c)} A robot with no rotational noise moves directly from one goal position to the next.
    \textbf{d)} A robot with rotational noise $\sigma$ moves in steps. For each one, it draws the step's random length from $\textit{Unif}(\frac12 b, \frac32 b)$ and its heading from the Gaussian distribution $\mathcal{N}(\theta^*(t), \sigma^2)$.
    \textbf{e, f)} At high densities, visible traffic jams occur in simulation and experiment. Green dots mark robots' goals. In \textbf{f)}, which shows a view of our robotic experiments from the overhead camera, red lights indicate that a robot's vision cone is currently occupied. Robot positions, headings, goals, and sensing cone edges are annotated. 
    }
    \label{fig:fig1_setting} 
\end{figure*}

In our noisy navigation setting, $n$ agents $a_1, a_2, ..., a_n$ begin at random start positions and move toward random goal positions, which are all drawn independently and uniformly from a 2D square of side-length $L$ (Fig.~\ref{fig:fig1_setting}\textbf{a}). When an agent reaches within tolerance $\epsilon$ of its goal, it receives a new random goal. 
Arena boundary conditions vary as specified below (periodic boundary conditions simplify calculations for theoretical analysis, but are not applicable to physical experiments). 

Agents move toward their goals in homing random walks, in which step lengths and heading angles are randomized. Animations of robot behavior are presented in SI Video 1. The heading angle $\theta$ at each step is drawn from a Gaussian distribution $\mathcal{N}(\theta^*(t), \sigma^2)$, whose mean is the optimal direction $\theta^*$ toward the goal at the beginning of the step (Fig.~\ref{fig:fig1_setting}\textbf{b}). (We note that the correctly invariant distribution for periodic variables is the von Mises-Fisher distribution~\cite{fisher1953dispersion}. For small variances, the difference between the von Mises-Fisher and Gaussian distributions is small. We work with a wrapped Gaussian.) Random step lengths are drawn from a uniform distribution in the interval $(\frac12 b, \frac32 b)$, where the constant $b$ represents average step length (if the agent does not stop for traffic during the step). Examples of homing walks with $\sigma = 0$ (no noise) and $\sigma = 0.75$ are shown in Fig.~\ref{fig:fig1_setting}\textbf{c}-\textbf{d}. 

\begin{figure*}
    \centering
    \includegraphics[width=0.8\linewidth]{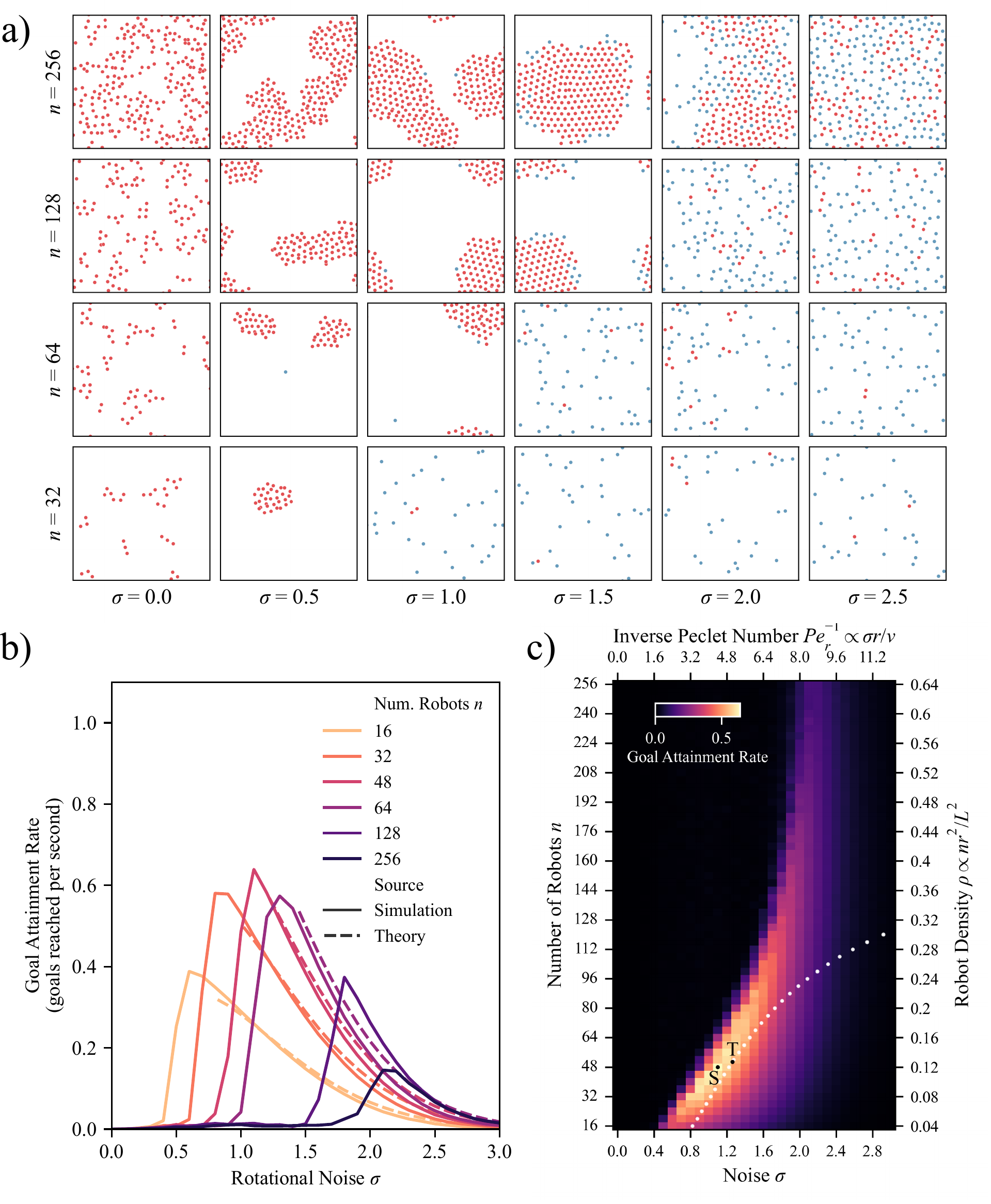}
    \caption{ \textbf{Simulations and theory.} 
    \textbf{a)} Jammed and dilute regimes. Dots show robot positions for various $(n, \sigma)$ after 8000 seconds of simulated time.  Color indicates whether robots are blocked (red) or free to move (blue). All subplots in this figure use periodic boundary conditions, $L = 40, r = 2, \gamma = \frac{2\pi}{3}, v = 0.5, b = 0.5, \epsilon = 0.6$. See Table~\ref{table:parameters} for more parameter details, and SI Video 2 for an animated version of this phase diagram.
    \textbf{b)} Solid lines plot simulation data for goal attainment rate $G$, the total goals reached per second by the entire swarm. Dashed lines show our theoretical approximations of $G$ for the dilute regime. The dashed lines end where our theory predicts the dilute regime ends. (The absence of dashed lines for $n \in \{128, 256\}$ reflects Eq.~\ref{eq:sigma_star} diverging at high densities; the theory remains valid in the regime of interest where the optimal parameters lie.) Goal attainment rate is computed over the last 25\% of the simulation time to capture steady state behavior. Simulations are run for 50 trials in the high-variance regime ($n \leq 128$ and $\sigma \leq 2.0$), and for 20 trials otherwise.
    \textbf{c)} Goal attainment across $(n, \sigma)$ space. The `S' dot marks the optimal $(n, \sigma)$ observed in simulations, the `T' dot marks the optimal $(n, \sigma)$ predicted by theory, and white dots mark the predicted optimal $\sigma$ predicted by theory for each $n$. 
    }
    \label{fig:fig2_theory}
\end{figure*}

Each agent has a forward sensing cone, a $\gamma = 120^{\circ}$ sector of a circle with radius $r$ (Fig.~\ref{fig:fig1_setting}\textbf{b}). When the sensing cone is unoccupied, agents move forward with speed $v$. When an agent senses a neighbor within the sensing cone, it ceases forward motion but can still rotate to the random angles drawn for new steps. The standard deviation $\sigma$ of the Gaussian angular noise is the same for all agents, and is the key parameter for tuning the noisiness of the team. The nondimensional parameters, which are also plotted in Fig.~\ref{fig:fig2_theory}\textbf{c}, are $\frac{r}{v} \sigma \propto Pe_r^{-1}$ (the inverse Peclet number) and $\frac{r^2}{L^2}n \propto \rho$ (density). Jams resulting from simulation and experiment are shown in Fig.~\ref{fig:fig1_setting}\textbf{e}-\textbf{f}. If the robot turns to an angle where its vision cone is free again, it resumes moving forward at the new angle. When $\sigma$ is higher and the motion is noisier, robots turn to a larger range of angles and are more likely to escape jams. 

We begin by observing simulations to understand the typical behavior of teams with different values of $n$ and $\sigma$. Simulation code is informed by the Stage simulator \cite{vaughan2008massively}. Agents proceed to their goals until they encounter another robot and stop moving forward. When there is no noise to unjam collisions ($\sigma = 0$), robots remain jammed in small clusters (leftmost column of Fig.~\ref{fig:fig2_theory}\textbf{a}). As $\sigma$ increases (moving toward the right in Fig.~\ref{fig:fig2_theory}\textbf{a}), the noise allows robots to escape some jam configurations. ``Easy" jams to escape usually contain few robots or have squiggly, irregular shapes; the persistent jams that remain are larger and rounder. As $\sigma$ increases farther, the steady state eventually becomes a single large jam containing essentially all the agents. When $\sigma$ is higher than a critical value $\sigma^*(n)$, the noise melts even this all-encompassing jam. Persistent jams no longer appear, and collisions are unjammed quickly by the high noise. We call the $\sigma < \sigma^*$ regime ``the jammed phase", and the $\sigma > \sigma^*$ regime where large jams do not form ``the dilute phase". The steady state goal attainment rate in the jammed phase is approximately 0 (left sides of solid curves in Fig.~\ref{fig:fig2_theory}\textbf{b}). In agreement with \cite{casiulis2021self}, we observe that noise levels just above $\sigma^*(n)$ maximize goal attainment for a given $n$ (peaks of solid curves in Fig.~\ref{fig:fig2_theory}\textbf{b}). Intuitively, once $\sigma$ is high enough to break up jams, adding more noise only reduces goal attainment by making travel paths more inefficient (right sides of solid curves in Fig.~\ref{fig:fig2_theory}\textbf{b}). The same information is presented as a heatmap in Fig.~\ref{fig:fig2_theory}\textbf{c}, where the sharp transition between the jammed and dilute phases is clearly visible. 

\section*{Analytic approximations}
With this qualitative understanding of the jammed and dilute phases, and the observation that goal attainment rate $G$ peaks near the boundary between the two, we next develop quantitative approximations of steady-state behavior in both phases. We then use this model to predict the optimal settings of $\sigma$ and $n$ that maximize $G$. 

We first approximate $G(n,\sigma)$ in the dilute phase. We then estimate the critical value of noise $\sigma^*(n)$ marking the edge of jamming for a team of $n$ robots. The optimal goal attainment rate for a team of size $n$ can then be estimated as $G(n, \sigma^*(n))$. Finally, we find the $n$ maximizing $G(n, \sigma^*(n); L, r, \gamma, b, v)$ by numerical optimization. This gives the predicted optimal team size for an arena of given side length and robots with given geometric parameters.

Our primary assumptions are that the system is at a steady state, the arena has periodic boundary conditions, step size $b$ is small, robots turn instantaneously, and (in the dilute regime where large jams do not form) all collisions are two-robot collisions. Derivation details, along with discussion of when and why approximation errors arise, are presented in SI section~\ref{section:si_derivations}.

\emph{Approximating $G$ in the dilute regime:}
The average initial straight-line distance between a robot and its random goal position is (see SI~\ref{section:si_goal_distance}, also in \cite{casiulis2021self}) \begin{equation}
\mathbb{E}(d) = \frac{\ln(3 + 2\sqrt{2}) + 2^{3/2}}{12}L \approx 0.3826L.
\label{eq:goal_distance}
\end{equation}

Though adding noise to an agent's motion helps it escape jams, it also makes its path to the goal less efficient. To capture this tradeoff, we consider the \textit{walk extension $\ext$}, the average ratio between the start-to-goal length of a homing random walk with noise $\sigma$ and a straight line. This quantity is also useful for polymer physics \cite{marantan2018mechanics} and for other studies of noisy searching processes, such as animal foraging \cite{hartman2024walk}. In the limit of small $b$ (see SI~\ref{section:si_extension}),
\begin{equation}
\ext \approx e^{\sigma^2/2}.
\label{eq:extension}
\end{equation}

In the absence of obstacles, the expected travel time $t_0$ to reach the goal is therefore
\begin{equation}
    \mathbb{E}(t_0) \approx \frac{0.3826L \cdot e^{\sigma^2/2}}{v}.
\label{eq:goal_travel_time}
\end{equation}
Next, we consider how many collisions $C$ a robot in the dilute regime experiences en route to a goal. 

We approximate the average collisions per goal as $[\text{robot density} \times \text{steps to reach goal} \times \text{vision cone area coverage per step}]$ (see SI~\ref{section:si_collision_frequency}):

\begin{equation}
\begin{split}
    \mathbb{E}(C) \approx \frac{n-1}{L^2} &\times \frac{0.3826 L \ext}{b} 
    \\ &\times \left[\frac{\gamma}{2\pi}\pi r^2 + 2\sqrt{2}br\sin(\frac{\gamma}{2})\right].
\end{split}
\label{eq:collisions_per_goal}
\end{equation}

To estimate the time $t_\textrm{jam}$ that robots spend jammed in each collision, we make the simplifying assumptions that all collisions are two-body and head-on, and that the jam ends when at least one robot turns enough to free its sensing cone (SI Fig.~\ref{fig:si_collision_duration}.) This occurs if a robot draws a random heading $\theta$ for which $|\theta - \theta^*| > \frac{\gamma}{2}$, the probability of which can be obtained by integrating over the Gaussian distribution of the rotational noise. Thus (see SI~\ref{section:si_two_body_escape})

\begin{equation}
    \mathbb{E}(t_\textrm{jam}) \approx \frac{b/v}{1 - \erf(\frac{\gamma}{2\sqrt{2}\sigma})^2}
\label{eq:collision_escape_time}
\end{equation}

The total time $t_g(n, \sigma)$ for a robot to reach one goal is then given by the sum of the time spent traveling and the time spent in jams:
$t_g(n, \sigma) \approx \mathbb{E}(t_0) + \mathbb{E}(C)\cdot \mathbb{E}(t_\textrm{jam})$.

Therefore, in the dilute regime where $\sigma > \sigma^*(n)$, the total goal attainment rate $G(n, \sigma)$ for the robot team is 
\begin{equation}
    G(n, \sigma) \approx \frac{n}{t_g(n,\sigma)}
\label{eq:swarm_goal_attainment}
\end{equation}

\emph{Approximating critical noise level $\sigma^*$:}
To solve for the noise level $\sigma^*$ at the edge of jamming for a given $n$, we approximate the expected time it takes for a robot to enter a jam from a random location, and the time to exit the jam. At the edge of jamming, these times will be equal.

Based on Fig.~\ref{fig:fig2_theory}\textbf{a}, we make the approximation that in the jammed regime at the edge of jamming, all robots except one are in one large circular jam. We approximate the robots as circles of radius $r/2$ and area $\pi r^2/4$; therefore the large jam has area $A_j \approx n\pi r^2/4$ and radius $r_j  \approx r\sqrt{n}/2$.  

We approximate the probability that a robot encounters the jam while traveling to a goal as $A_j/L^2 = n\pi r^2/4L^2$. Thus the robot will reach about $4L^2/n\pi r^2$ goals before getting stuck in the jam. Using Eq.~\ref{eq:goal_travel_time}, the time this takes is (see SI~\ref{section:si_jam_entry_time})

\begin{equation}
    \mathbb{E}(\text{jam entry time}) \approx \frac{1}{v} \frac{4 \cdot 0.3826 \cdot L^3 \cdot \ext}{n \pi r^2}
\label{eq:jam_entry_time}
\end{equation}

To estimate the jam escape time, we assume: (1) in an average case, the robot needs to travel $1/4$ of the jam perimeter to escape the jam; (2) the robot is in an ``average'' position (shown in Fig.~\ref{fig:si_large_jam_position}) where it needs to turn $\frac{\pi}{4} + \frac{\gamma}{2}$ to free its sensing cone, which occurs with probability $\frac{1}{2} - \frac12 \erf(\frac{x}{\sqrt{2}\sigma})$ per update step; (3) the $1/4$-perimeter distance the robot needs to travel is scaled by both $\ext$ and $\frac{1}{\frac{1}{2} - \frac12 \erf(\frac{x}{\sqrt{2}\sigma})}$. This overestimates the difficulty of exiting the jam, so the final calculated $\sigma^*$ will be an overestimate (see SI~\ref{section:si_jam_escape_time}). The jam exit time can then be approximated as
\begin{equation}
    \mathbb{E}(\text{jam exit time}) \approx \frac{1}{2}\frac{\pi r \sqrt{n}\ext}{v(1-\erf((\frac{\pi}{4} + \frac{\gamma}{2})\frac{1}{\sqrt{2}\sigma}))}.
\label{eq:jam_exit_time}
\end{equation}
By setting Eqs.~\ref{eq:jam_entry_time} and \ref{eq:jam_exit_time} equal, we solve for the optimal noise level
\begin{equation}
    \sigma^*(N) \approx (\frac{\pi}{4} + \frac{\gamma}{2})\frac{1}{\sqrt{2}}(\erf^{-1}(1 - \frac18 \frac{\pi^2 r^3 n^{3/2}}{0.3826L^3}))^{-1}
\label{eq:sigma_star}
\end{equation}

For a given value of $n$, the optimal goal attainment rate is then $G(n, \sigma^*(n)) = n/t_g(n, \sigma^*(n))$. 
To find the optimal set of values $(n, \sigma)$ that maximizes goal attainment rate overall, we numerically maximize $G(n, \sigma^*(n))$ as a function of $n$. 

\emph{Approximation accuracy:}
The parameters selected are close to the optimal ones observed in simulations, as shown in Fig.~\ref{fig:fig2_theory}\textbf{c}. In Fig.~\ref{fig:fig2_theory}\textbf{b} and \textbf{c}, we see that predictions for attainment rate and $\sigma^*(n)$ diverge as $n$ grows large. This aligns with several assumptions we made that break down at high $n$ or $\sigma$. For example, we assumed that all collisions are two-body collisions; the multibody interactions that we neglected grow more frequent at higher densities. Using this assumption to obtain analytical results is the current state of the art. It is also employed, for example, by \cite{bacik2023lane}. See SI section~\ref{section:si_derivations} for detailed discussion of error sources.

For further testing, SI Fig.~\ref{fig:si_abcdef_sweep} compares our predicted optimal $(n, \sigma)$ with the optimal parameters observed in additional simulation settings varying one of $r$, $\gamma$, or $b$. The approximations are accurate except when the optimal $n$ is large (in settings that make the sensing cone smaller, so more agents can effectively share the space). This again reflects the approximation's divergence at high densities.

\section*{Robotic experiments}

\begin{figure*}
    \centering
    \includegraphics[width=1\linewidth]{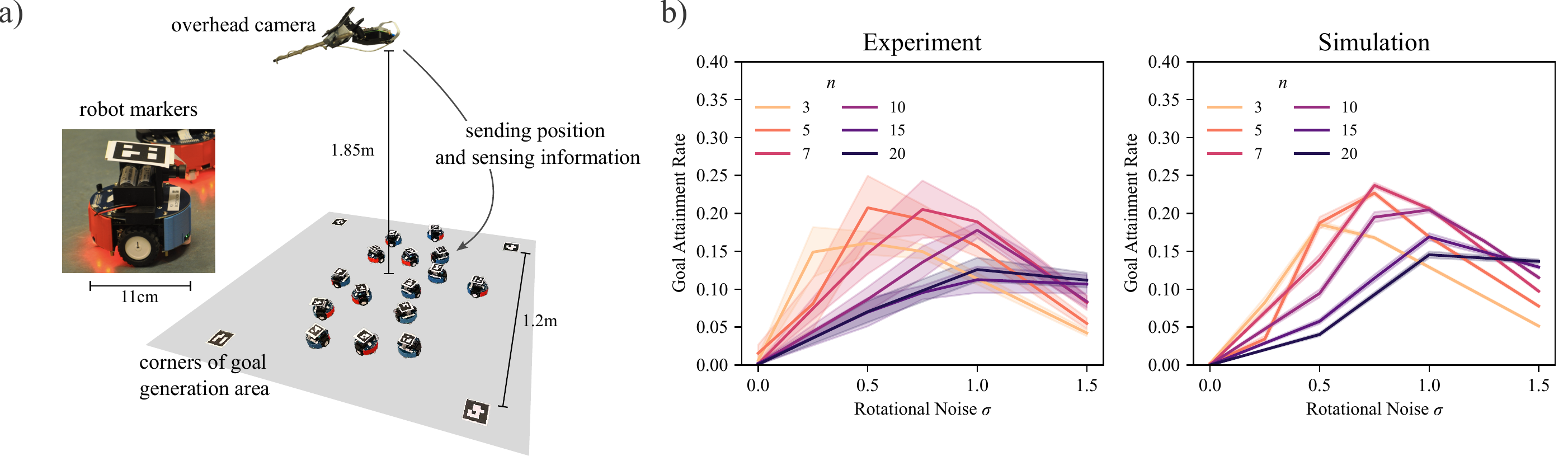}
    \caption{ \textbf{Robotic experiments. }
    \textbf{a)} Individual robot and schematic of experimental setup.
    \textbf{b)} Goal attainment rate for experiments vs. simulations with matching parameters. Experimental data have 6 trials per point; simulation data have 100 trials per point. Data includes the last 75\% of 300-second trials. Shaded error bands show standard error. See Table~\ref{table:parameters} for parameter details and SI Video 3 for experimental footage.
    }
    \label{fig:fig3_experiments}
\end{figure*}

The results we presented so far use ideal assumptions, such as the sensing cone being fully reliable and the robot speed being fully controllable. To evaluate the behavior of physically instantiated systems, we repeat the above experiments on teams of Alphabot2-Pi robots matching those used in \cite{alonso2024single} to study single-file motion. The physical experiments investigate system behavior under more realistic conditions, in contrast to idealized assumptions used by the theory and simulations, such as periodic boundary conditions and small step sizes. For the relatively simple Alphabots, robot speed is neither constant nor homogeneous, and sensing neither perfect nor instantaneous. 

We couple the robot experiments with a new set of simulations whose parameters and conditions match those of the physical experiments. While robots are programmed to have $r = 0.2$m and $\gamma = \frac{2\pi}{3}$, in practice they stop at a closer distance than $0.2$m due to communication lag. We compute an ``effective radius" ($r = 0.156$m) based on experimental data to use as the sensing range in the matching simulations. Similarly, we use experimental data to estimate robots' effective linear and turning speeds for each $(n, \sigma)$ setting, and we use the corresponding values in the simulations. See the Materials and Methods section and SI Section~\ref{section:si_experiments} for more experimental details.

The physical and simulated experiments show qualitative and quantitative agreement. Qualitatively, goal attainment rate curves are nonmonotonic in both density and noise level, with peaks moving to higher noise levels as density increases (Fig.~\ref{fig:fig3_experiments}\textbf{b}). Quantitatively, the simulations and experiments agree on the $\sigma$ where goal attainment peaks (Fig.~\ref{fig:si_optimal_sigma_match}).

\section*{Adding intelligent behavior to agents}
\begin{figure*}
    \centering
    \includegraphics[width=1\linewidth]{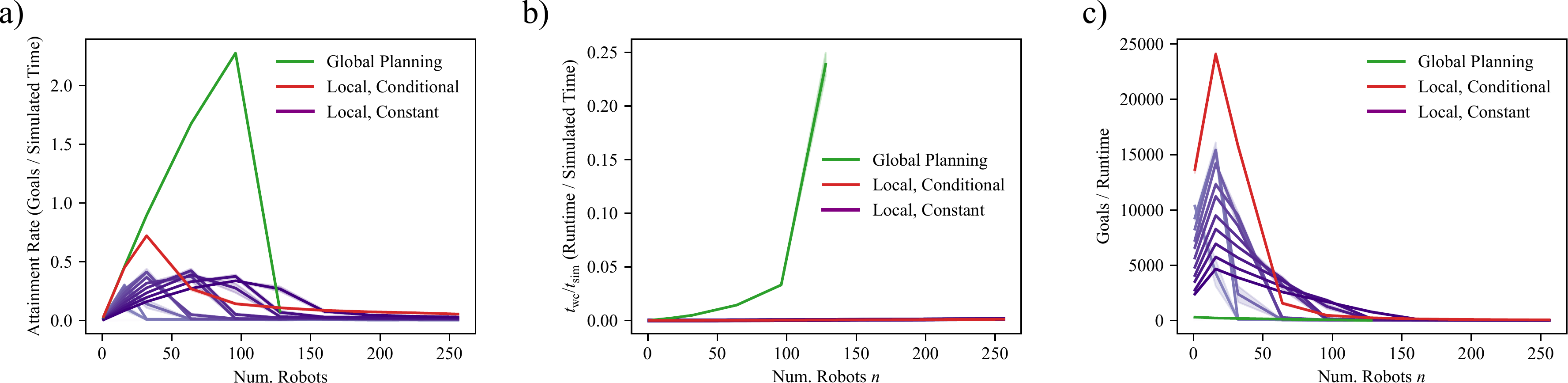}
    \caption{ \textbf{More intelligent controllers.}
    In all subplots, the purple lines for local, constant noise represent different values of $\sigma$ ranging from $0.5$ to $1.5$. Shaded error bands show standard error. Data contains 20 trials per point. See Table~\ref{table:parameters} for more parameter details.
    \textbf{a)} Comparing the attainment rate of different kinds of noisy navigation controllers. See SI Video 4 for animated examples of the different controllers.
    \textbf{b)} Comparing the ratio of runtime to simulated time for local and global controllers.
    \textbf{c)} Comparing the goals reached per wall-clock time of local and global controllers.
    }
    \label{fig:fig_intelligence}
\end{figure*}

Having seen what we can achieve in the absence of intelligence, a natural question is how much we can gain with more intelligent agents. We compare the noisy motion controller with constant Gaussian noise against two more intelligent control methods. The first method is noisy motion with conditional noise, another distributed approach where robots react to local sensing. Robots using conditional noise move directly toward the goal unless they detect a neighbor in their sensing cone. When a robot's sensing cone is occupied, it draws a uniform random angle for its next step. Conditional noise is still distributed and depends entirely on local sensing, but adds ``one bit" of intelligence. The second method we test is Cooperative A* \cite{silver2005cooperative}, a greedy global planning algorithm for finding non-colliding paths on a graph. In the graph, each node represents occupying a cell at a certain time. The algorithm looks for agent paths to their goals in individual, sequential searches; Cooperative A* does not perform a joint search that returns all agents' paths at once. 
See SI Section~\ref{section:si_global_planner} for implementation details. 

\newcommand{\twc}[0]{t_{\text{wc}}}
\newcommand{\tsim}[0]{t_{\text{sim}}}
We see in Fig.~\ref{fig:fig_intelligence}\textbf{a} that the simple Gaussian noise controller studied earlier reaches about half the attainment of the conditional noise method. In turn, the conditional noise method reaches about 30\% of the global A* planner's maximum attainment. At low densities, the attainment of the conditional noise tracks closely with that of the global planner. 
However, the global planner is computationally expensive, and the wall-clock time $\twc$ needed to simulate an in-simulation time $\tsim$ increases rapidly, which can be seen by plotting $\twc/\tsim$ in Fig.~\ref{fig:fig_intelligence}\textbf{b}. All simulations are run on a 2021 M1 Macbook Pro.
Because the local sensing methods are simulated with finer timesteps, their runtimes are in fact overestimated compared to the global planner. Furthermore, the local mechanism is inherently parallelizable while the global planner relies on sequential planning. To assess how efficiently the different controllers utilize computation, in Fig.~\ref{fig:fig_intelligence}\textbf{c} we plot goals reached per $\twc$. The local planner with conditional noise reaches the highest goal attainment per unit of computation: 80 times that of the maximum attained with A* global planning and 1.6 times the maximum attained with constant Gaussian noise. 

\section*{Discussion}
Inspired by the problem of steric constraints that can hinder task execution by agents, we studied how noisy motion can help simple agents reach individual goal locations in a crowded environment. Our integrative approach combines theory, simulation, and experiments to corroborate key system behaviors from multiple perspectives.

We found that a noisy local approach attained high efficiency relative to how much computation it used. Furthermore, the stochasticity of the Gaussian noise controller allowed us to perform approximate optimizations over global parameters (agent density and noise level). This suggests that looking for other systems where stochasticity produces predictable steady-state behavior may help us optimize the parameters for more sophisticated multibody collectives. A natural further direction is to study how maximum swarm capability increases as intelligence and computational complexity is added to a collective. How much complexity can be shed without losing significant performance? 

Our robotics setting shares some parallels with the dynamics of active matter~\cite{tailleur2022active}, especially in the context of an effective field theory in the hydrodynamic limit associated with long wavelength, slow dynamics of the agent density and its variations in space-time. Natural variables for this are robot density and speed, which are both functions of space and time. Directed motion towards a target can be described with a non-local gradient term that serves to orient robots towards their individual targets, subject to keeping the total number of robots constant. The jam formation process can potentially be modeled with a continuum framework to describe the likelihood of different sized jams appearing, while jam dissolution is a function of diffusive processes at play. Finally, adding intelligence is tantamount to including additional non-local terms that describe the locations (or probability distributions) of agents. Working towards such a theory is likely to involve generalizations of paradigms such as the formation and breakup of traffic jams~\cite{lighthill1955kinematic,newell1993simplified} and motility-induced phase separation~\cite{cates2015motility}, but with additional terms arising from global considerations, questions worthy of future study.

\textbf{Data Availability.} Simulation code and data from the robotic experiments are available at \href{https://doi.org/10.5281/zenodo.17954636}{https://doi.org/10.5281/zenodo.17954636} \cite{lucy_liu_2025_17954636}.

\bibliography{bib}

\onecolumngrid  
\appendix

\renewcommand{\theequation}{S\arabic{equation}}
\renewcommand{\thefigure}{S\arabic{figure}}
\renewcommand{\thetable}{S\arabic{table}}
\renewcommand{\thesection}{S\arabic{section}}
\renewcommand{\thesubsection}{\arabic{subsection}}
\renewcommand{\thealgorithm}{S\arabic{algorithm}}

\setcounter{equation}{0}
\setcounter{figure}{0}
\setcounter{table}{0}
\setcounter{section}{0}
\setcounter{algorithm}{0}

\section{Local Sensing Simulation \label{section:si_local_simulation}}
Here we present additional details regarding the local sensing simulation. When the boundaries are not periodic, there are no boundary conditions. Robot goals are limited to the $L \times L$ goal generation area, but robots with high enough noise to stray from their goals can exit that area. Simulations for the constant noise controller use small step lengths to better match theoretical assumptions (see Table~\ref{table:parameters} for details), while simulations for the conditional noise controller use larger step lengths so that the noisy steps are long enough to be consequential.

If a robot uses the homing random walk controller, its reflexive decision making proceeds as shown in Algorithm~\ref{alg:constant_noise_update}. (Note that Algorithm~\ref{alg:constant_noise_update} describes a robot which turns instantaneously. If this is not the case, the robot first turns to its new travel angle at the start of each step; time spent turning is not counted in the step duration.)

\begin{algorithm}[H]
\caption{Constant Noise Agent Update}
\begin{algorithmic}[1]
\If{agent has reached within $\epsilon$ of goal}
    \State draw new goal
    \State end current step
\EndIf
\If{starting a new random walk step}
    \State draw random duration of new step from Unif$(\frac{1}{2}\frac{b}{v}, \frac{3}{2}\frac{b}{v})$
    \State draw random new \texttt{travel angle} from $\mathcal{N}(\texttt{angle to goal}, \sigma^2)$
    \State \texttt{heading} $\gets$ \texttt{travel angle}
\EndIf

\If{vision cone blocked}
    \State \texttt{fwd speed} $\gets$ $0$
\Else 
    \State \texttt{fwd speed} $\gets$ $v$
\EndIf

\end{algorithmic}
\label{alg:constant_noise_update}
\end{algorithm}

The conditional noise controller is a slightly modified version of the constant noise controller, and is presented in Algorithm~\ref{alg:conditional_noise_update} below.

\begin{algorithm}[H]
\caption{Conditional Noise Agent Update}
\begin{algorithmic}[1]
\If{agent has reached within $\epsilon$ of goal}
    \State draw new goal
    \State end current step
\EndIf
\If{starting a new random walk step}
    \State draw random duration of new step from Unif$(\frac{1}{2}\frac{b}{v}, \frac{3}{2}\frac{b}{v})$

    \If{vision cone blocked}
        \State draw uniform random new \texttt{travel angle}
    \Else 
        \State \texttt{travel angle} $\gets$ \texttt{angle to goal}
    \EndIf
    \State \texttt{heading} $\gets$ \texttt{travel angle}
\EndIf

\If{vision cone blocked}
    \State \texttt{fwd speed} $\gets$ $0$
\Else 
    \State \texttt{fwd speed} $\gets$ $v$
\EndIf

\end{algorithmic}
\label{alg:conditional_noise_update}
\end{algorithm}

Though robots stop when they detect a neighbor within their vision cone of radius $r$, it is still possible for the distance between two robots to be less than $r$. This can be seen in some close-together robot pairs in Fig.~\ref{fig:fig2_theory}\textbf{a}. For example, two robots moving nearly parallel to each other can become very close without blocking each other's sensing cones. Alternatively, it is possible for robot $a_1$ to draw a random step angle that causes it to move further into robot $a_2$'s sensing cone, all while $a_1$'s sensing cone remains free.

\section{Derivation Details \label{section:si_derivations}}

\subsection{Random Distance on Torus \label{section:si_goal_distance}}
\begin{figure}
    \centering
    \includegraphics[width=0.9\linewidth]{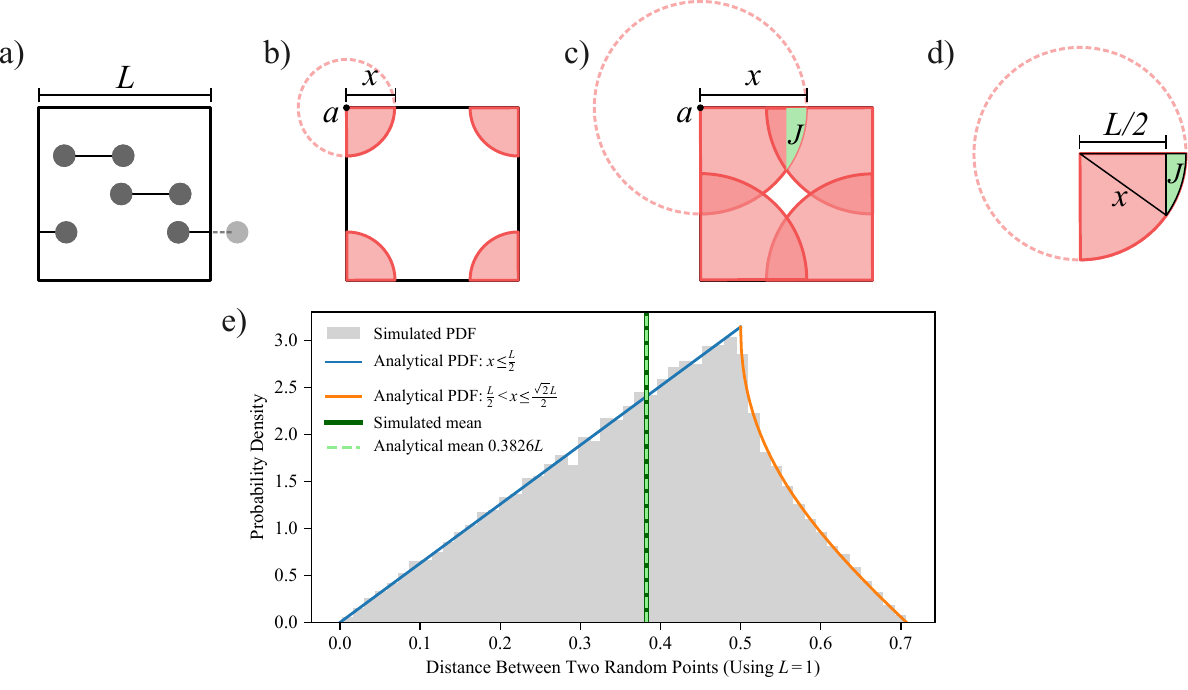}
    \caption{
    \textbf{Statistics of distances on a torus.}
    \textbf{a)} A square of side length $L$ that has periodic boundary conditions, which we also call a torus. Several pairs of points and the shortest paths between them are drawn. For the bottom pair, the point on the left has the same position as the faded point, but wrapped across the periodic boundaries.
    \textbf{b)} When $x \leq \frac{L}{2}$, the area within $x$ of a point $a$ (shaded) is the area of a circle.
    \textbf{c)} When $\frac{L}{2} <  x \leq \frac{\sqrt{2}}{2}L$, parts of the wrapped circle begin to overlap. To compute the the area within $x$ of a point $a$, we can subtract the double-counted regions of area $J$.
    \textbf{d)} The area $J$ can be computed by finding the areas of a circle sector and a triangle.
    \textbf{e)} Comparing analytical results from main text Eqn.~\ref{eq:goal_distance} to simulations using a torus with $L = 1$ and 50000 randomly generated distances.  
    }
    \label{fig:si_torus_distances}
\end{figure}

    Here we derive main text Eqn.~\ref{eq:goal_distance}, which states the average distance between a robot and its random goal (when the $L\times L$ arena has periodic boundary conditions). To do so, we will calculate the expected distance between two random points on a square with periodic boundary conditions, which we also call a torus. Let $A, B$ be two random 2D points drawn uniformly and independently from $[0, L]^2$. Their corresponding PDFs are constant functions over the square: if $a$ is a possible location for $A$ and $b$ is a possible location for $B$, then $f_A(a) = f_B(b) = \frac{1}{L^2}$. We are interested in the random distance $X = \Vert A - B\Vert$. Since $A$ and $B$ lie on a topological torus, this distance refers to the length of the shortest possible path between them, accounting for how positions wrap across the boundaries. Some examples of shortest distances are shown in Fig.~\ref{fig:si_torus_distances}\textbf{a}. We will compute the CDF of $X$, use the CDF to obtain the PDF, and finally integrate over the PDF to find the expected value $\mathbb{E}(X)$, the average distance between $A$ and $B$. Since robots' start and goal locations are also uniform random points, $\mathbb{E}(X)$ will equal a robot's average distance to its random goal. An alternate version of this derivation which finds the PDF first is presented in \cite{casiulis2021self}.

\textbf{CDF.} We start with the CDF definition and marginalize over possible locations for $A$ using Law of Total Probability. 
\begin{align}
    P(X \leq x) 
    &= \int_{[0,L]^2} P\left(X \leq x \quad|\quad A=a\right)f_A(a) da \notag
    \\ &= \int_{[0,L]^2} P\left(X \leq x \quad|\quad A = a\right)\frac{1}{L^2} da \notag
    \\ &= \frac{1}{L^2}\int_{[0,L]^2} P\left(\Vert a-B\Vert  < x\right)da
\end{align}

We've now turned our statistics problem into a very visual geometry problem! $P(\Vert a-B\Vert  < x)$ for a known $a$ is the fraction of the torus within $x$ of point $a$. For a typical shape without periodic boundary conditions, this fraction depends on how close to the edge $a$ is, but on a torus, which has no boundary, the notion of distance from the edge is meaningless. Thus, $P(\Vert a-B\Vert  < x) = \frac{\text{torus area within $x$ of $a$}}{L^2} $ is a constant with no dependence on the position of $a$, and it can be moved outside of the integral. Outside the integral, $a$ can be considered to be any fixed point in the torus; for convenience we set $a=(0, L)$, the point at the top left corner.
\begin{align}
    P(X \leq x)  &= \frac{1}{L^2} \frac{\text{torus area within $x$ of $a$}}{L^2} \int_{[0,L]^2} da \notag
    \\ &= \frac{1}{L^2} \frac{\text{torus area within $x$ of $a$}}{L^2} L^2 \notag
    \\ &= \frac{\text{torus area within $x$ of $a$}}{L^2}
\end{align}

When $x \leq \frac{L}{2}$, the area within $x$ of $a$ is the entire radius-$x$ circle of area $\pi x^2$, because the circle can wrap across the periodic boundaries. This is shown in Fig.~\ref{fig:si_torus_distances}\textbf{b}.

When $x > \frac{L}{2}$, computing $P(\Vert a - B\Vert  < x)$ using only the circle area is no longer sufficient, because the wrapped circle arcs begin to overlap (depicted in Fig.~\ref{fig:si_torus_distances}\textbf{c}.) To correct for this when computing the area of the torus within $x$ of $a$, we subtract the overlapping area, which is $8J$ for the entire torus. ($J$ is shaded in green in Fig.~\ref{fig:si_torus_distances}\textbf{c}.) Now we can write the CDF for this case as

\begin{equation}
    P(X \leq x) = \frac{\pi x^2 - 8J}{L^2}
\end{equation}

We compute $J$ by finding the area of the circle sector outlined in black in Fig.~\ref{fig:si_torus_distances}\textbf{d} and subtracting the area of the triangle drawn within it. 

\begin{equation}
    J = \frac{x^2}{2}\arccos(\frac{L}{2x}) - \frac{L}{4}\sqrt{x^2 - \frac{L^2}{4}}
\end{equation}

 When $x > \frac{\sqrt{2}L}{2}$, half the length of the ``diagonal" of the torus, the entire torus will be within $x$ of $a$. Putting these cases together, we have found the CDF of $X$.

The CDF is thus
\begin{align}
    P(X \leq x) 
    &= 
\begin{cases}
    \frac{\pi x^2}{L^2} & x \leq \frac{L}{2} \\
    \frac{1}{L^2}\left(\pi x^2 - 4x^2\arccos(\frac{L}{2x}) + 2L\sqrt{x^2 - \frac{L^2}{4}}\right) & \frac{L}{2} < x \leq  \frac{\sqrt2 L}{2}\\
    1 & \frac{\sqrt2 L}{2} < x
\end{cases}
\label{eq:si_torus_cdf}
\end{align}

\textbf{PDF.} We find the PDF of random distances on a torus by taking the derivative of the CDF (Eqn.~\ref{eq:si_torus_cdf}) with respect to $x$:
\begin{align}
f_X(x) = 
\begin{cases}
    \frac{2\pi x}{L^2} & x \leq \frac{L}{2} \\
    \frac{1}{L^2} \left(2{\pi}x-8x\arccos\left(\frac{L}{2x}\right)\right)) & x > \frac{L}{2} \\
    0 & x > \frac{\sqrt2 L}{2}
\end{cases}
\label{eq:si_torus_pdf}
\end{align}

\textbf{Expected Value.} Finally, we compute the expected value by plugging Eqn.~\ref{eq:si_torus_pdf} into the integral definition. This recovers main text Eqn.~\ref{eq:goal_distance}.
 \begin{align*}
     \mathbb{E}(X) &= \int_0^{\sqrt2 L / 2} x \cdot f_X(x)dx
     \\ &= \int_0^{L/2} \frac{x^2}{L^2}2\pi dx + \int_{L/2}^{\sqrt2 L / 2} \frac{x^2}{L^2}\left( 2\pi - 8\arccos\left(\frac{L}{2x}\right) \right) dx
     \\ &= \frac{\pi L}{12} + \frac{L}{12}\left(\ln(3 + 2\sqrt{2}) + 2^{3/2} - \pi \right)
     \\ &= \frac{\ln(3 + 2\sqrt{2}) + 2^{3/2}}{12}L
     \\ &\approx 0.3826L
 \end{align*}

\textbf{Comparison to Simulation}

We verify the  main text Eqn.~\ref{eq:goal_distance} approximation by comparing it to simulations, which is shown in Fig.~\ref{fig:si_torus_distances}\textbf{e}.

\subsection{Inefficiency of Noisy Paths \label{section:si_extension}}

\begin{figure}
    \centering
    \includegraphics[width=0.75\linewidth]{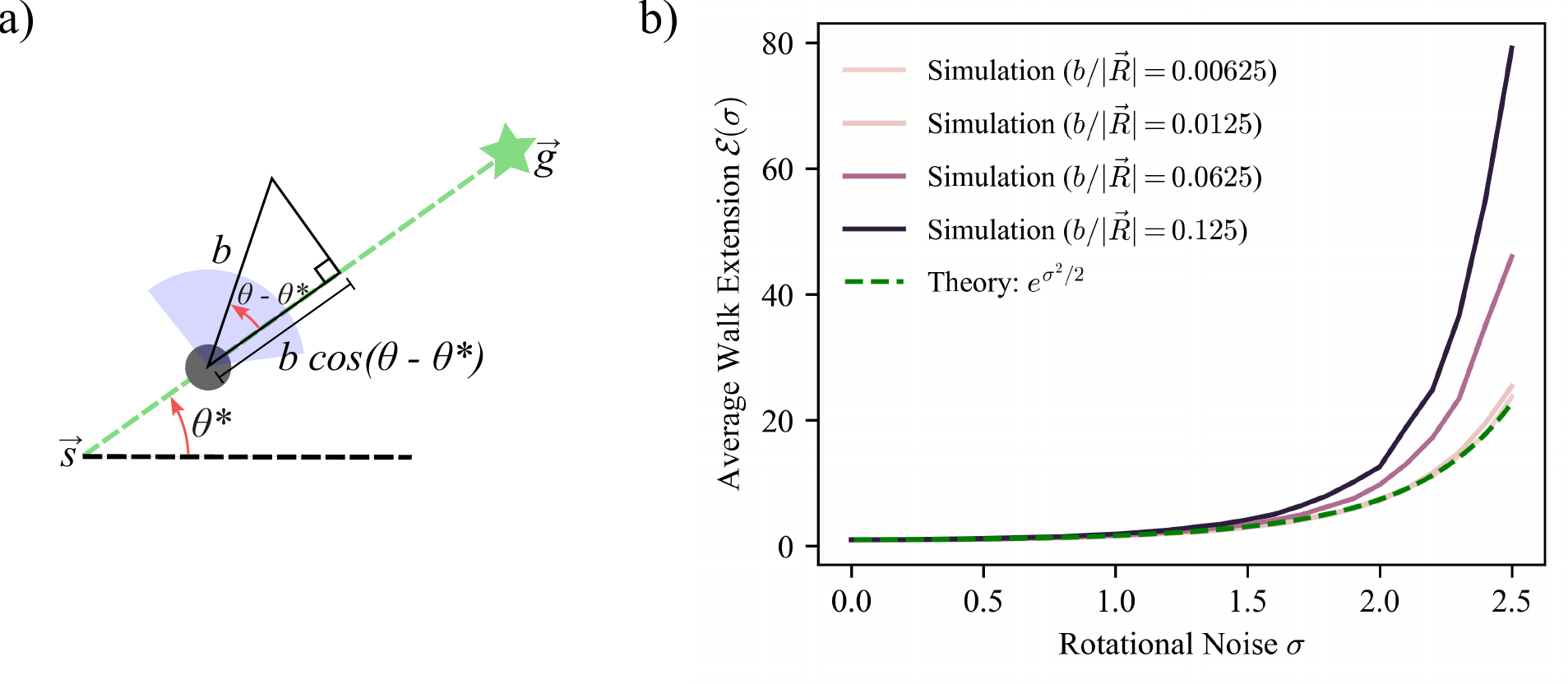}
    \caption{\textbf{Noisy walk extension approximation.}
    \textbf{a)} Diagram of variables referenced in derivation setup.  The new length-$b$ step is the hypotenuse of the right triangle drawn in black. It is depicted to be large instead of in the small $b$ limit to improve visualization.
    \textbf{b)} Approximation accuracy of main text Eqn.~\ref{eq:extension} for homing random walk lengths of different noise levels $\sigma$ and step sizes $b$. The theory diverges for large step sizes and high noise. Simulations are run for 500 trials per point and use $\Vert \Vec{R}\Vert  = 8$ and $\epsilon = 0.5$.}
    \label{fig:si_extension}
\end{figure}

Here, we derive the \textit{walk extension} $\mathcal{E}(\sigma)$ presented in Equation~\ref{eq:extension}, the expected ratio between the length of a noisy walk from start to goal and the length of a direct line from start to goal. Our key assumption is that the robots' step size $b$ is small enough that walks deviate relatively little from the straight-line path to the goal, so that $\theta^*$ remains essentially constant. This allows us to treat the distribution of the rotational noise as $\mathcal{N}(\theta^*, \sigma^2)$.

Let $\Vec{s}$ and $\Vec{g}$ be the start and goal positions, and let $\Vec{R} = \Vec{g} - \Vec{s}$ so that $\Vert \Vec{R}\Vert $ is the L2 distance to the goal. The optimal heading toward the goal at time 0 is $\theta^*(0)$. During a noisy homing walk toward $\Vec{g}$ in the small step-size limit $\frac{b}{\Vert \Vec{R}\Vert } \rightarrow 0$, the walk position strays very little from the direct line to the goal (drawn in green in Fig.~\ref{fig:si_extension}\textbf{a}) because the steps are so small. Then the optimal direction toward the goal remains essentially constant, and $\theta^*(t) \approx \theta^*(0)$ for all times $t$ until $\Vec{g}$ is reached. This is no longer the case if steps are large, as a large step orthogonal to $\theta^*(0)$ can cause $\theta^*(t)$ to change significantly. The approximation also breaks down if noise is very high, which also increases the likelihood of large deviations from the direct line to the goal. 

Then the random walk heading angles $\theta$ follow the Gaussian distribution $\mathcal{N}(\theta^*, \sigma^2)$. Consider a new step $\Vec{v}$ of the random walk with average length $b$ and heading $\theta \sim \mathcal{N}(\theta^*, \sigma^2)$. In Fig.~\ref{fig:si_extension}\textbf{a}, $\Vec{v}$ would be the hypotenuse of the right triangle drawn in black. A noiseless, direct walk to the goal will take $\frac{\Vert \Vec{R\Vert }}{b}$ steps on average, whereas a homing walk with noise $\sigma$ will take $\frac{\Vert \Vec{R}\Vert }{\mathbb{E} \Vert \operatorname{proj}_{\Vec{R}}\Vec{v}\Vert }$ steps on average. Here, $\Vert \operatorname{proj}_{\Vec{R}}\Vec{v}\Vert  = b\cos(\theta  - \theta*)$ is the progress the random step makes in the direction of the goal. Then the extension can be written as the ratio between the average steps taken by the noisy walk and the direct walk:
\begin{equation}
    \ext = \frac{b}{\mathbb{E} \Vert \operatorname{proj}_{\Vec{R}}\Vec{v}\Vert }
\label{eq:si_noisy_direct_ratio}
\end{equation}

Setting $\theta^* = 0$ without loss of generality and integrating over the support of the Gaussian distribution,
\begin{align}
    \mathbb{E} \Vert \operatorname{proj}_{\Vec{R}}\Vec{v}\Vert  &= b\int_{-\infty}^{\infty} \cos(\theta) p(\theta) d\theta &\text{(Expected value definition)} \notag
    \\ &= b\int_{-\infty}^{\infty} \cos(\theta) \frac{1}{\sqrt{2\pi \sigma^2}} e^{-\theta^2/(2\sigma^2)} d\theta &\text{(Gaussian PDF)} \notag
    \\ &= b\frac{1}{\sqrt{2\pi \sigma^2}} \int_{-\infty}^{\infty} \cos(\theta) e^{-\theta^2/(2\sigma^2)} d\theta \notag
    \\ &= b\frac{1}{\sqrt{2\pi \sigma^2}} \cdot [\sqrt{2\pi \sigma^2} \cdot e^{-\sigma^2/2} \cdot \lim_{x \rightarrow \infty} \operatorname{Re}(\erf(\frac{x + i\sigma^2}{\sqrt{2}\sigma}))] &\text{(See Eqn.~\ref{eq:si_erf_cos_integral} below)} \notag
    \\ &= b e^{-\sigma^2/2}
\label{eq:si_extension_projection}
\end{align}

because $\lim_{x \rightarrow \infty} \operatorname{Re}(\erf(\frac{x + i\sigma^2}{\sqrt{2}\sigma})) = 1$. 

Plugging Eqn.~\ref{eq:si_extension_projection} into Eqn.~\ref{eq:si_noisy_direct_ratio}, we recover main text Eqn.~\ref{eq:extension}:
$$\ext = \frac{b}{\mathbb{E} \Vert \operatorname{proj}_{\Vec{R}}\Vec{v}\Vert } = \frac{b}{b e^{-\sigma^2/2}} = e^{\sigma^2/2}.$$

\textit{Additional identities. }In deriving Eqn.~\ref{eq:si_extension_projection}, we used the following identity with $a = 1, b = \frac{1}{2\sigma^2}$:

\begin{equation}
    \int_{-z}^{z} \cos(ax) e^{-bx^2} \, dx  = \sqrt{\frac{\pi}{b}} e^{ - \frac{a^2}{4b}} \operatorname{Re}\left[\erf(\sqrt{b}z + \frac{ai}{2\sqrt{b}}) \right].
\label{eq:si_erf_cos_integral}
\end{equation}

\textit{Proof of Eqn.~\ref{eq:si_erf_cos_integral}.}
\begin{align*}
    \int_{-z}^{z} \cos(ax) e^{-bx^2} \, dx &=     \frac{1}{2} \int_{-z}^{z} \left[ e^{iax} + e^{-iax} \right] e^{-bx^2} \, dx &\text{(Exponential formula for cosine)} \\
    &= \frac{1}{2} \left[ \int_{-z}^{z} e^{iax-bx^2} \, dx + \int_{-z}^{z} e^{-iax-bx^2} \, dx \right]
    \\ &= \frac{1}{2} \left[ \int_{-z}^{z} e^{-b(x - \frac{ai}{2b})^2 - \frac{a^2}{4b}} \, dx + \int_{-z}^{z} e^{-b(x + \frac{ai}{2b})^2 - \frac{a^2}{4b}} \, dx \right] &\text{(Complete the square)}
    \\ &= \frac{1}{2} e^{ - \frac{a^2}{4b}}\left[ \int_{-z}^{z} e^{-b(x - \frac{ai}{2b})^2} \, dx + \int_{-z}^{z} e^{-b(x + \frac{ai}{2b})^2} \, dx \right]
    \\ &= \frac{1}{2} e^{ - \frac{a^2}{4b}} \cdot \frac{1}{\sqrt{b}} \cdot \frac{\sqrt{\pi}}{2} \cdot 2 \left[ \erf(\sqrt{b}(z - \frac{ai}{2b})) + \erf(\sqrt{b}(z + \frac{ai}{2b})) \right] &\text{(See Eqn.~\ref{eq:si_erf_integral} below)}
    \\ &= \sqrt{\frac{\pi}{b}}\frac{1}{2}e^{ - \frac{a^2}{4b}}\left[ 2\operatorname{Re}(\erf(\sqrt{b}z + \frac{ai}{2\sqrt{b}})) \right]  &(\text{Because } \erf(\overline{z}) = \overline{\erf(z)})
    \\ &= \sqrt{\frac{\pi}{b}} e^{ - \frac{a^2}{4b}} \operatorname{Re}\left[\erf(\sqrt{b}z + \frac{ai}{2\sqrt{b}}) \right] \qed
\end{align*}

Our proof for Eqn.~\ref{eq:si_erf_cos_integral} in turn relies on the identity
\begin{equation}
\int_{-z}^{z} e^{-b(t+c)^2}dt = \frac{1}{\sqrt{b}} \frac{\sqrt{\pi}}{2} \left[ \erf(\sqrt{b}(z+c)) + \erf(\sqrt{b}(z-c)) \right].
\label{eq:si_erf_integral}
\end{equation}

\textit{Proof of Eqn.~\ref{eq:si_erf_integral}.}
\begin{align*}
    \int_{-z}^{z} e^{-b(t+c)^2}dt &= \int_{\sqrt{b(c-z)}}^{\sqrt{b}(c+z)} e^{-u^2}\frac{1}{\sqrt{b}}du &\text{(Change of variables)}
    \\ &= \frac{1}{\sqrt{b}} \left [ \int_{0}^{\sqrt{b}(c+z)} e^{-u^2}du + \int_{\sqrt{b(c-z)}}^{0} e^{-u^2}du \right ]
    \\ &= \frac{1}{\sqrt{b}} \frac{\sqrt{\pi}}{2} \left[ \erf(\sqrt{b}(c+z)) - \erf(\sqrt{b}(c-z)) \right] &\text{(Error function definition)}
    \\ &= \frac{1}{\sqrt{b}} \frac{\sqrt{\pi}}{2} \left[ \erf(\sqrt{b}(z+c)) + \erf(\sqrt{b}(z-c)) \right] \qed &(\erf(-z) = -\erf(z))
\end{align*}

We validate the Equation~\ref{eq:extension} approximation in Fig.~\ref{fig:si_extension}\textbf{b} by comparing its predictions to the lengths of single-agent noisy walks taken in simulation. Our approximation is accurate when $b$ is small relative to the goal distance and $\sigma$ is not large. This agrees with the earlier discussion of cases where our $\theta^*(t) \approx \theta^*(0)$ assumption breaks down. In the final step of a walk, robots may traverse only part of the step before reaching within $\epsilon$ of the goal and ending the walk. To measure the partial step length traversed, we use line-circle collision detection code from \cite{thompson_collisiondetection}.

\subsection{Dilute Regime: Collision Frequency \label{section:si_collision_frequency}}

\begin{figure}
    \centering
    \includegraphics[width=\linewidth]{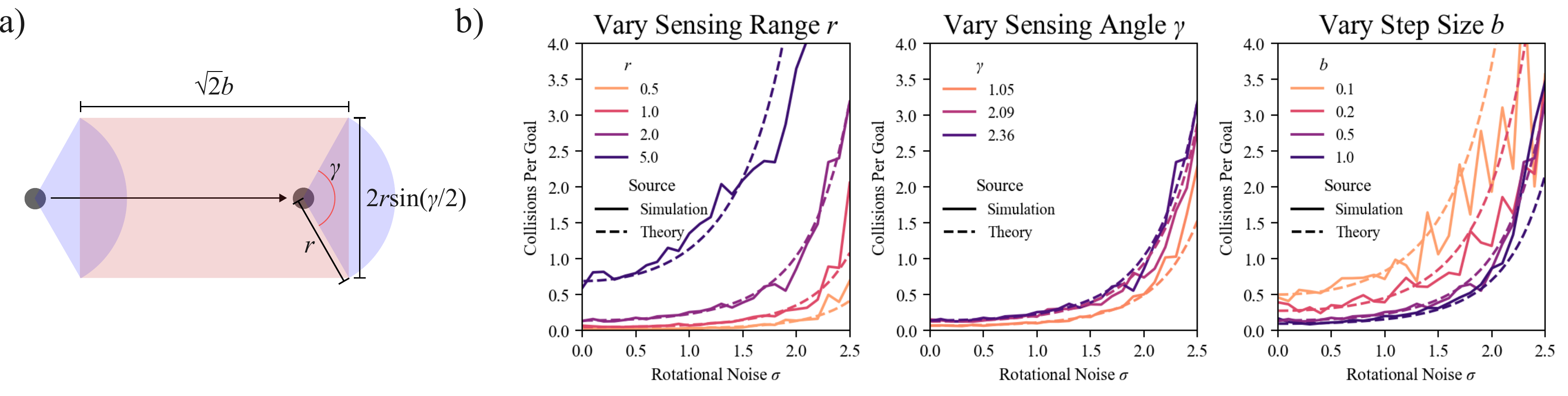}
    \caption{\textbf{Collision frequency approximation.}
    \textbf{a)} Diagram showing the sensing cone (blue shading) and elongation area (pink shading) of a step referenced in the collision frequency derivation. The entire shaded region represents the area covered by a sensing cone during a step.
    \textbf{b)} Collision frequency approximation compared to dilute simulations where $n=2$. In each panel, $r$, $\gamma$, or $b$ is varied while other variables are held fixed. The parameter values (unless otherwise specified) are $r=2.0, \gamma= \frac{2\pi}{3}\approx 2.36, b=0.5$, the same values used in main text Fig.~\ref{fig:fig2_theory}. Simulations are run for 100 trials of 10000 seconds each. All other parameters match those used in Fig.~\ref{fig:fig2_theory}. 
    }
    \label{fig:si_collision_frequency}
\end{figure}

Now that we understand the behavior of a robot traveling to its goal in isolation, we begin making estimates that account for interactions with other robots. In the main text, these are Equations~\ref{eq:collisions_per_goal}-\ref{eq:jam_exit_time}. First, we work in the dilute regime depicted in the right-side columns of Fig.~\ref{fig:fig2_theory}\textbf{a}, where noise is high enough that large jams do not persist. In the dilute regime, we assume that robots are distributed uniformly through space. (This assumption requires the boundary conditions to be periodic. Without periodic boundary conditions, robots' shortest paths to the goal are more likely to cross through the center of the arena than its edges, causing density to be nonuniform.)

In main text Equation~\ref{eq:collisions_per_goal}, we approximated the ratio of collisions encountered to goals reached as 

$$\frac{\text{num. collisions}}{\text{num. goals}} = \frac{N-1}{L^2} \cdot \frac{0.3826 L \ext}{b} \cdot \left[\frac{\gamma}{2\pi}\pi r^2 + 2\sqrt{2}br\sin(\frac{\gamma}{2})\right].$$

Here, $\frac{N-1}{L^2}$ is the density of other robots in the arena,  $\frac{0.3826 L \ext}{b} $ is the average number of step segments the robot takes to reach each goal, and $\left[\frac{\gamma}{2\pi}\pi r^2 + 2\sqrt{2}br\sin(\frac{\gamma}{2})\right]$ approximates the area covered by an agent's sensing cone during a step. 

We break the sensing cone coverage area into the base area $\frac{\gamma}{2\pi}\pi r^2$ of the sensing cone itself (shaded in blue in Fig.~\ref{fig:si_collision_frequency}\textbf{a}) and an \textit{elongation area} (shaded in pink in Fig.~\ref{fig:si_collision_frequency}\textbf{a}) describing the additional area covered due to the distance traveled during a step. We approximate the elongation area as $\sqrt{2}b \times 2r\sin(\frac{\gamma}{2})$ based on the extra area an agent views as it moves forward. The $\sqrt{2}$ accounts for an agent's mean speed relative to another agent moving in a random direction, and the $2r\sin(\frac{\gamma}{2})$ approximates the ``width" of the sensing cone. Our approximation ignores finer nuances of the area coverage geometry, such as how vision cone area coverage may overlap between neighboring steps in a walk.

In Fig.~\ref{fig:si_collision_frequency}\textbf{b}, we compare our theory to dilute simulations where $N=2$. A collision is logged when an agent's sensing cone becomes blocked. The collision ends at the next timestep where the sensing cone is free again. The collision frequency approximation is accurate unless $\sigma$ is high and $b$ is large or small. When $b$ is large, error arises from the $\ext$ estimate breaking down. When $b$ is small, the error comes from neighboring steps in a noisy walk covering overlapping areas. Our approximation did not account for this and assumed area coverage of different steps was independent of each other.

\subsection{Dilute Regime: Two-Body Jam Escape Time\label{section:si_two_body_escape}}
\begin{figure}
    \centering
    \includegraphics[width=0.75\linewidth]{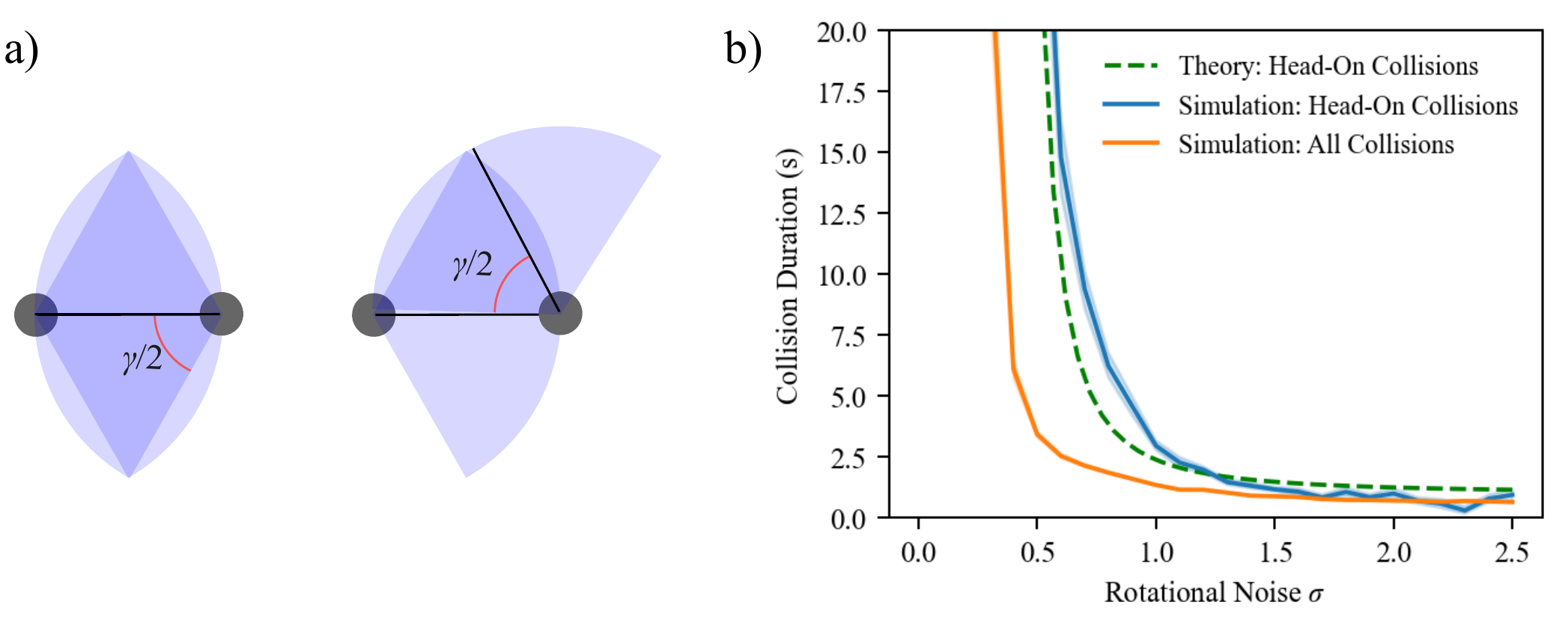}
    \caption{\textbf{Collision duration approximation.} \textbf{a)} Left: two robots in a head-on collision. Right: one robot turns just over $\frac{\gamma}{2}$, the minimum turn angle needed to free its sensing cone. (A neighbor is detected only if the neighbor's center point is in the robot's sensing cone.)
    \textbf{b)} Main text Eqn.~\ref{eq:collision_escape_time} approximation compared to simulations where two robots with $\gamma = \frac{2\pi}{3}$ collide.}
    \label{fig:si_collision_duration}
\end{figure}

In Equation~\ref{eq:collision_escape_time}, we approximated the jam escape time $\mathbb{E}(t_\textrm{jam})$ in the dilute regime. Our simplifying assumptions were that all collisions were two-body and head-on, and that jams ended when at least one robot turned enough to free its sensing cone. By ``head-on", we mean that if each robot faced its optimal heading $\theta^*$, the two robots would lie on the center lines of each other's sensing cones. 

In a head-on collision, at least one robot needs to turn an angle of magnitude $\frac{\gamma}{2}$ or more to free its sensing cone, which is illustrated in Fig.~\ref{fig:si_collision_duration}\textbf{a}. The probability that a single robot doesn't turn enough (that is, it turns to an angle $\theta$ such that $|\theta - \theta^*| \leq \frac{\gamma}{2}$) is $\erf(\frac{\gamma}{2\sqrt{2}\sigma})$:

\begin{align}
    P(|\theta - \theta^*| \leq \frac{\gamma}{2}) &= \int_{-\gamma/2}^{\gamma/2} \frac{1}{\sqrt{2\pi \sigma^2}} e^{-x^2/(2\sigma^2)}dx &\text{(Gaussian PDF)} \notag
    \\ &= \frac{1}{\sqrt{2\pi \sigma^2}} \int_{-\gamma/2}^{\gamma/2} e^{-x^2/(2\sigma^2)}dx \notag
    \\ &= \frac{1}{\sqrt{2\pi \sigma^2}} \cdot \sqrt{2\sigma^2 \pi} \cdot \frac12 \cdot 2 \cdot \erf\left(\frac{\gamma}{2}\frac{1}{\sqrt{2}\sigma}\right) &\text{(Applying~\ref{eq:si_erf_integral})} \notag
    \\ &= \erf \left(\frac{\gamma}{2\sqrt{2}\sigma}\right)
\label{eq:si_p_doesnt_turn_enough}
\end{align}

It follows that if a pair of new angles are drawn for the robots, the probability that at least one robot turns enough and frees the jam is:

\begin{equation}
    P(\text{two-body jam freed}) = 1 - \erf(\frac{\gamma}{2\sqrt{2}\sigma})^2
\label{eq:si_two_robot_p_good_turn}
\end{equation}

(Random steps may have different lengths, but since the average length is fixed, over time two robots should draw the same number of new steps and angles.) The random angles drawn in different steps are independent, so we expect $\frac{1}{1 - \erf(\frac{\gamma}{2\sqrt{2}\sigma})^2}$ rounds of angles to be drawn until the robots are freed. Since a step on average has length $b$ and takes $b/v$ seconds, the expected time that this takes is (this recovers main text Eqn.~\ref{eq:collision_escape_time}):

$$\mathbb{E}(t_\textrm{jam}) \approx \frac{b/v}{1 - \erf(\frac{\gamma}{2\sqrt{2}\sigma})^2}$$

Head-on jams are the most difficult to exit. By assuming all of the jams are head-on, we overestimate the average exit time. In Fig.~\ref{fig:si_collision_duration}\textbf{b}, we compare our theory to measurements of collision duration in an $n = 2$ dilute setting. Our estimate $\mathbb{E}(t_\textrm{jam})$ matches the data for head-on collisions, but overestimates the collision duration for a general collision which may not be head on. For each collision detected in Fig.~\ref{fig:si_collision_duration}\textbf{b}, we compute a value $h$ measuring how far the collision is from being head-on. Consider a collision between two agents $a_i$ and $a_i$ whose optimal headings toward their respective goals are $\theta^*_i$ and $\theta^*_j$. We define $\alpha_{ij}$ as the angle of the vector pointing from $a_i$ to $a_j$ and correspondingly define $\alpha_{ji}$, so that $|\theta^*_i - \alpha_{ij}|$ measures the difference between the angles from $a_i$ to its goal and from $a_i$ to $a_j$. Then 

\begin{equation}
    h(\theta^*_i, \theta^*_j, \alpha_{ij}, \alpha_{ji}) = \sqrt{(\theta^*_i - \alpha_{ij})^2 + (\theta^*_j - \alpha_{ji})^2}
\end{equation}

The data for simulated head-on collisions in Fig.~\ref{fig:si_collision_duration}\textbf{b} is taken from the same collision dataset used for Fig.~\ref{fig:si_collision_frequency}\textbf{b}, but includes only collisions with $\gamma = \frac{2\pi}{3}$ and $h < 0.2$.

By assuming the jams are all two-robot jams, we have ignored multibody collisions, which take longer to exit from and occur more frequently at higher densities. The increasing approximation error for higher densities is visible in Fig.~\ref{fig:fig2_theory}\textbf{b}.

\subsection{Jammed Regime: Time to Enter Jam\label{section:si_jam_entry_time}}

The approximation presented in Equation~\ref{eq:jam_entry_time} estimates the probability of travel to a new goal causing a robot to encounter the large jam as $\frac{\text{jam area}}{\text{arena area}}$. This estimate does not account for cases where a goal does not lie inside the jam, but the path to the goal must cross the jam. 

\subsection{Jammed Regime: Time to Escape Jam\label{section:si_jam_escape_time}}
To approximate the escape time from a large jam for main text Eqn.~\ref{eq:jam_exit_time}, we use strong simplifying assumptions. In many two-robot jams, a single high-noise step forward often suffices to disperse the jam and allow both robots to resume traveling to their goals. However, if a robot needs to pass a large jam containing numerous robots, it may need many lucky, unblocked steps until the path to the goal is clear.

\begin{figure}
    \centering
\includegraphics[width=0.5\linewidth]{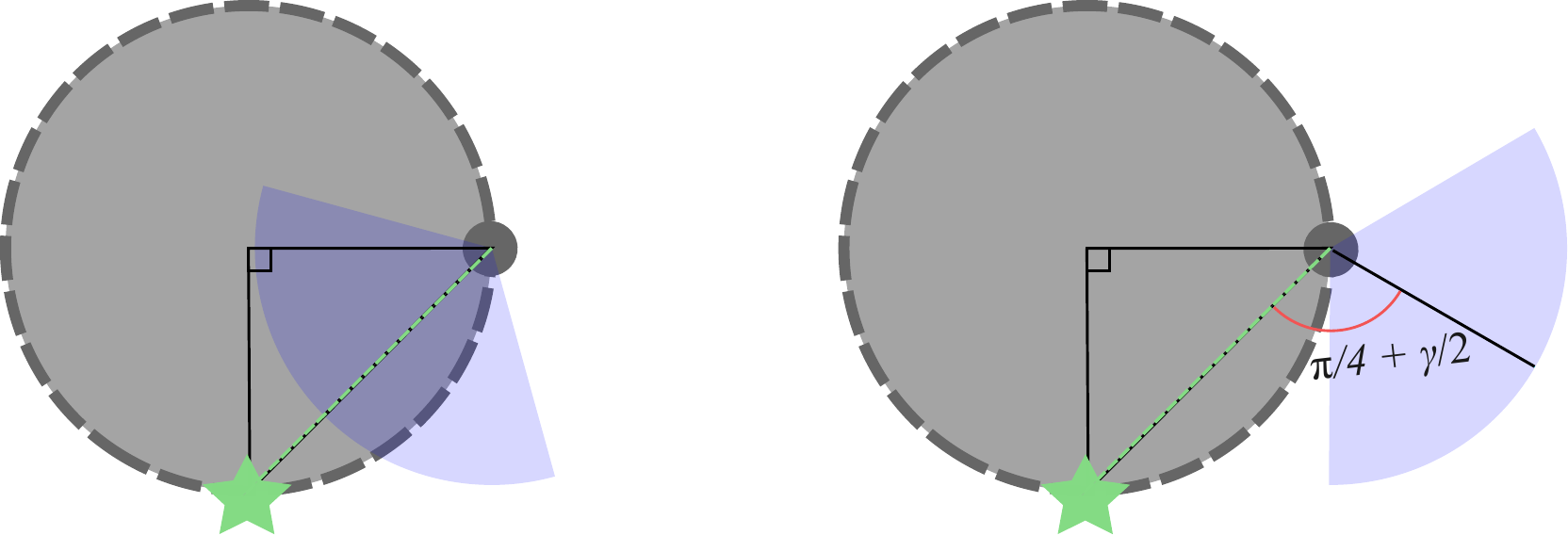}
    \caption{\textbf{Escape from a large jam.} The large jam is approximated as a circle (dashed outline). Left: Robot positioning in the average case described in text, where the robot (smaller gray circle) is a quarter-perimeter away from its goal (green star). Right: Robot turns the minimum angle away from the jam needed to free its sensing cone.}
    \label{fig:si_large_jam_position}
\end{figure}

For the estimate we give in Equation~\ref{eq:jam_exit_time}, we approximate the jam as a solid circular obstacle of radius $r_j = r\sqrt{n}/2$ and perimeter $\pi r \sqrt{n}$. Right after a robot enters the large jam, it is positioned on the jam perimeter. The closest possible point on the jam perimeter is a distance of 0 away, and the farthest possible point on the perimeter is a half-perimeter walk away. We assume that in an average case, the robot needs to travel $\frac14$ of the jam perimeter to escape the jam either by reaching its goal and concluding the homing walk, or by reaching a position where it can continue moving toward the goal without being obstructed by the jam. (We do not account for cases where the goal lies deep within the jam interior.) If there is noise, but no traffic encounters, this takes about $\frac14 \frac{\pi r \sqrt{n} \ext}{b}$ random walk steps.

To approximate the probability that a new random step angle frees the robot's sensing cone, we assume that the robot is always in a position a quarter-perimeter away from the goal, where it can free its sensing cone and take a step if it turns more than $\frac{\pi}{4} + \frac{\gamma}{2}$ away from the jam. A schematic of the robot position is shown in Fig.~\ref{fig:si_large_jam_position}. In this model, the probability of turning so the sensing cone is freed is approximately $P(\theta - \theta^* >  \frac{\pi}{4} + \frac{\gamma}{2})$. 

Because we know $\theta$ is a Gaussian with standard deviation $\sigma$, we use the Gaussian CDF to see that $P(\theta - \theta^* 
\leq x) = \frac12(1 + \erf(\frac{x}{\sqrt{2}\sigma}))$. Then 
\begin{equation}
P(\theta - \theta^* > x) = 1 - P(\theta - \theta^* 
\leq x) = \frac{1}{2} - \frac12 \erf(\frac{x}{\sqrt{2}\sigma}).\end{equation}

Then in expectation, the robot is unblocked once every $\frac{1}{\frac{1}{2} - \frac12 \erf(\frac{\frac{\pi}{4} + \frac{\gamma}{2}}{\sqrt{2}\sigma})}$ steps. Since one step takes on average $\frac{b}{v}$ seconds, our final estimate for the exit time, given in main text Equation~\ref{eq:jam_exit_time}, is
\begin{align*}
    \mathbb{E}(\text{jam exit time}) &\approx \frac{b}{v} \cdot \frac14 \frac{\pi r \sqrt{n} \ext}{b} \cdot \frac{1}{\frac{1}{2} - \frac12 \erf(\frac{\frac{\pi}{4} + \frac{\gamma}{2}}{\sqrt{2}\sigma})} 
    \\ &= \frac{1}{2}\frac{\pi r \sqrt{n}\ext}{v(1-\erf(\frac{\frac{\pi}{4} + \frac{\gamma}{2}}{\sqrt{2}\sigma}))}
\end{align*}

This estimate introduces several sources of error. For example, we scale the quarter-perimeter distance the robot travels by both $\ext$ (accounts for the noisy walk) and $1/P(\theta - \theta^* > \frac{\pi}{4} + \frac{\gamma}{2})$ (accounts for steps where the travel direction is blocked by the jam). These two effects are not independent, so scaling by both overcounts the difficulty they add to the robot's travel. 

Another source of error, which also applies to Eq.~\ref{eq:si_two_robot_p_good_turn}, is how when we use $P(\theta - \theta^* > \frac{\pi}{4} + \frac{\gamma}{2})$, we do not account for the periodicity of angles. For instance, if $\theta - \theta^* = 2\pi$, their difference is greater than $\pi$ even though $\theta$ and $\theta^*$ are the exact same angle. The error associated with the angular wrapping grows as $\sigma$ increases. 

To take a step unobstructed by the large jam, a robot could turn away from the jam (counterclockwise for the robot in Fig.~\ref{fig:si_large_jam_position}), or so far in the other direction (clockwise for the robot in Fig.~\ref{fig:si_large_jam_position}) that the vision cone is freed again. Our estimate does not account for this latter case. When $\sigma$ is low, it occurs rarely and error is minor. However, when $\sigma$ is high, the resulting error causes the $\sigma^*$ estimate to diverge, as seen in the black dots diverging from the empirical $\sigma^*$ value in Fig.~\ref{fig:fig2_theory}\textbf{c}. 

To account for the extra case, one can add a term corresponding to large turns toward the jam to our estimated probability of the sensing cone being freed. This modification is not used in any main text results, but we discuss it in the SI to check our understanding of the approximation error. The modification replaces $P(\theta - \theta^* >  \frac{\pi}{4} + \frac{\gamma}{2})$ with $P(\theta - \theta^* >  \frac{\pi}{4} + \frac{\gamma}{2}) + P(\theta - \theta^* <  \frac{3\pi}{4} + \frac{\gamma}{2})$. The rest of the jam exit time estimate proceeds as before, but solving for $\sigma^*$ is no longer analytically tractable. Using a numerical solve to find $\sigma^*$ instead, we verify that when this change is made, the $\sigma^*$ estimate no longer diverges at high $n$ as the estimate in Fig.~\ref{fig:fig2_theory}\textbf{c} does.

\subsection{Approximation Accuracy}

\begin{figure}
    \centering
\includegraphics[width=0.75\linewidth]{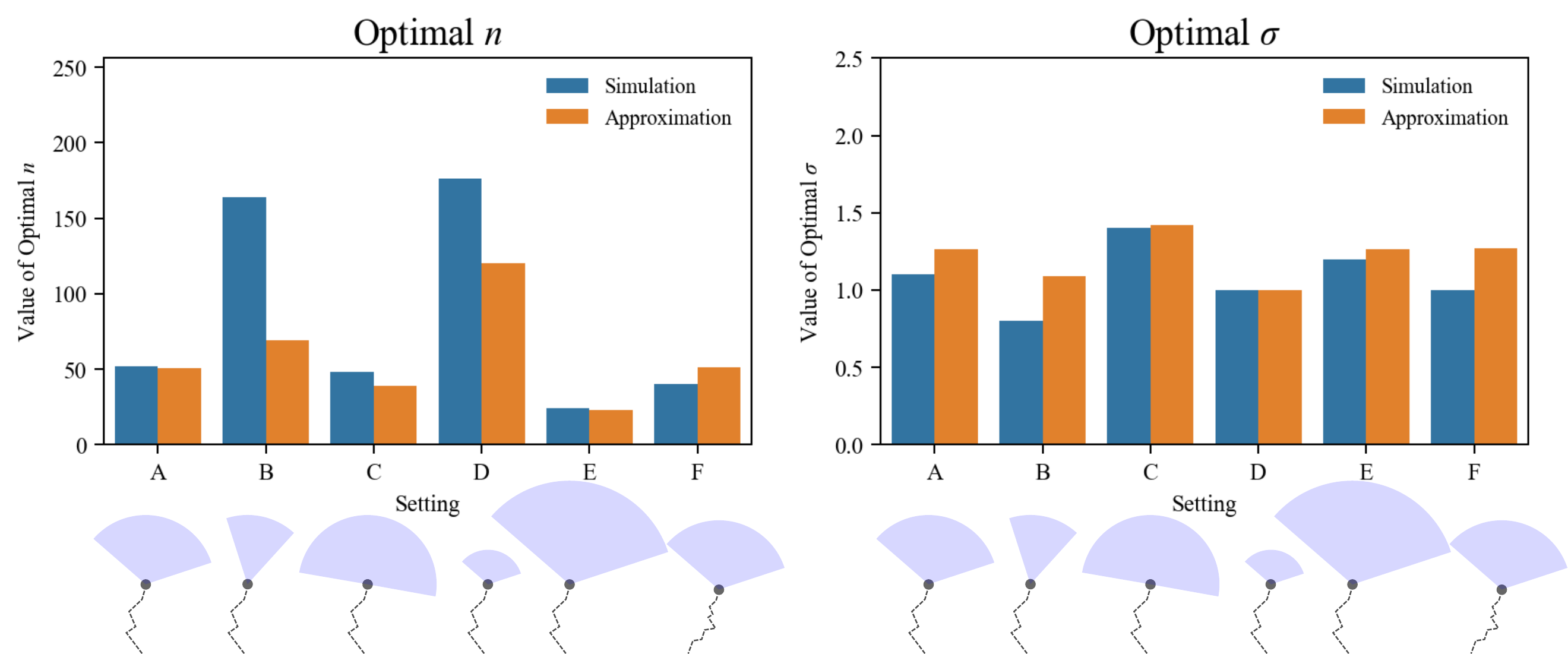}
    \caption{\textbf{Comparing the observed and predicted optimal team size and noise level for five different parameter settings.} A: same base setting as in Fig.~\ref{fig:fig2_theory}; B: smaller sensing angle $\gamma = \frac{\pi}{3}$; C: larger $\gamma = \pi$; D: smaller sensing range $r = 1$; E: larger $r = 3$; F: smaller step size $b = 0.25$.}
    \label{fig:si_abcdef_sweep}
\end{figure}

In the main text, our results for approximation accuracy used fixed values for $\gamma, r, b$. In Fig.~\ref{fig:si_abcdef_sweep}, we assess approximation accuracy for several settings where these parameters vary. Setting A matches the base setting shown in the rest of Fig.~\ref{fig:fig2_theory}. The other settings vary one of sensing cone angle, sensing cone length, or agent step size while keeping the other parameters the same. We see the greatest error when the optimal $n$ is large (in settings that make the sensing cone smaller and allow more agents to effectively share the space), reflecting the attainment approximation's inaccuracy at high densities and noise levels. This aligns with several error sources we discussed where our approximation breaks down in the high $n$, high $\sigma$ regimes.

\section{Experiments \label{section:si_experiments}}

\subsection{Experimental Setup}
The off-the-shelf Alphabots are outfitted with 3D printed battery holders and bumpers (blue bumper on the front and red bumper on the back). For the overhead camera, we use an ArduCam 64MP Hawkeye camera connected to a Raspberry Pi 5 at $2312\times 1736$ resolution. An MQTT broker facilitates communication between the overhead camera and the robots, which are also controlled by Raspberry Pis. Data from the MQTT broker is logged in an InfluxDB database. Since robot speeds vary as batteries deplete, trials are run in an order that cycles through noise levels for a given team size.

There are some differences between the boundary conditions in the experiments and the simulations. When nonperiodic boundary conditions are used in the simulations, the boundaries are completely free. All goals are generated within the $L \times L$ arena and robots home in toward their goals, but in noisier walks robots can also move outside the  $L \times L$ arena. In the experiments, a physical boundary made of wooden beams keeps robots from moving more than about 15-20cm outside of the goal generation area (whose corners are marked by ArUco markers). Robots bump into the wooden boundary until they draw a random angle that causes them to turn away. 

The experiments have an issue with neighbor detection when robots are outside of the $L \times L$ goal generation area. These robots are not detected as neighbors, and other robots do not stop for them until they physically collide. Robots spend $<3\%$ of their time in this outer edge region. Of this time, the outer-edge robots are within $r$ of other robots $<5\%$ of the time. Thus the detection issue impacts fewer than $0.15\%$ of observations, and we believe it has a negligible effect on results.

Another new element is stalling. The robots move at relatively low speeds so the camera can accurately capture their positions. Sometimes robots stall, failing to overcome friction and move, even when their motors activate as intended. When the camera detects stall events, robots briefly use higher motor speeds to make a shaking motion in an attempt to exit the stall. Robot speed is not homogeneous among robots, even though all robots use the same hardware and software. Robot speeds also vary as batteries deplete, so trials are run in an order that cycles through noise levels for a given team size. 

\subsection{Matching Simulations}
We can specify duty cycles that control the power of the robot motors, but cannot set an exact desired speed for the robot's forward or turning motion. Similarly, due to the time that elapses between  robots reaching a position and receiving the position update from the camera, it is challenging to precisely program each robot's sensing cone radius. To run simulations matching our robotic experiments, we extract estimates of turning speed, forward speed, and sensing cone radius from the experimental data. 

We obtain a robot's speed in between two position updates by dividing the position difference between the updates by the time difference between the updates. To estimate turning speed, we use only data from when robots are turning to their next heading, finding that robots turn at 1.039 rad/s on average (across all settings). To estimate forward speed, we use only data from when robots have finished turning and are not blocked, finding that robots move forward at 0.1 m/s on average (across all settings). For the simulations used in the main text, we compute a separate forward speed and turn speed for each $(n, \sigma)$ setting. 

We also extract robots' approximate effective vision cone radius from the experimental data. Though robots are programmed to have a vision cone with radius $r = 0.2$m, in practice they often stop at a smaller distance. This is because a robot only stops after the overhead camera detects a neighbor in its vision cone and sends that information to the robot. The physical robots can also move slightly while turning in place, so even blocked robots do not stay in precisely the same location. First, we observe that at the end of $20$-robot experimental trials, the nearest observed in-cone neighbor for blocked robots is on average 0.1425m. As discussed in Section~\ref{section:si_local_simulation}, even with perfect and instantaneous sensing like in the simulation, robots can reach closer than $r$ from their nearest neighbor. We observe from simulations where $r = 0.2$ that at trial end, the nearest in-cone neighbor for blocked robots is 91\% of the programmed radius. Then accounting for this scaling, the effective radius we compute for the experimental data is 0.1564m.

\begin{figure}
    \centering
    \includegraphics[width=0.5\linewidth]{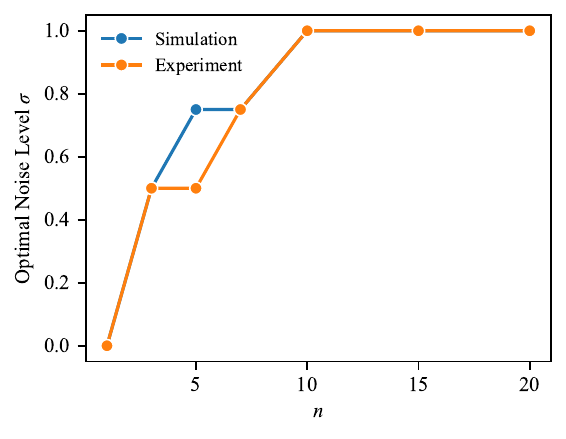}
    \caption{\textbf{Comparing the optimal noise level in experiments and matching simulations.} This plot is obtained from the data shown in main text Fig.~\ref{fig:fig3_experiments}\textbf{b}.}
    \label{fig:si_optimal_sigma_match}
\end{figure}

To assess how well the experiments and matching simulations match, we compute the optimal noise level from the $\sigma$ values tested for each $n$ (based on averages from the goal attainment data presented in Fig.~\ref{fig:fig3_experiments}\textbf{b}). We display the results in Fig.~\ref{fig:si_optimal_sigma_match} and observe agreement for all points except the $n=5$ point. For the $n=5$ case, the experiment's $\sigma^*$ value is $0.5$ while the simulation's $\sigma^*$ value is $0.75$. We did not test values in between $0.5$ and $0.75$, and the true $\sigma^*$ value is likely somewhere in between.

\section{Global Planning Simulation \label{section:si_global_planner}}
To measure the efficiency of a global planning procedure, we implement Cooperative A*. Cooperative A* \cite{silver2005cooperative} is a greedy algorithm for finding non-colliding paths on a graph. In the graph, each node represents occupying a cell at a certain time. Agents look for paths to the goal in individual, sequential searches; Cooperative A* does not perform a joint search that returns all agents' paths at once. When agents plan a path, they can access all existing plans other agents have made. A pseudocode outline is presented in Algorithm~\ref{alg:global_planner_update} below.

\begin{algorithm}[H]
\caption{Global Planner Update}
\begin{algorithmic}[1]

\For{each agent}
    \If{agent has reached goal goal}
        \State assign a new random goal
    \EndIf
\EndFor

\For{each agent needing a new plan}
    \State compute new plan using A* search based on shared global ledger
    \State log new plan in shared global ledger
\EndFor

\For{each agent}
    \State execute next step of agent's plan
\EndFor

\end{algorithmic}
\label{alg:global_planner_update}
\end{algorithm}
We discretize the $L\times L$ arena into a grid of cells. An agent occupying a cell is positioned at the cell's center point. When an agent receives a goal, it uses the A* graph search algorithm to find a path to its goal cell. A path to the goal has the form $((t_1, x_1, y_1), (t_2, x_2, y_2), ...)$ where $t_i$ marks the time of step $i$, and $x_i, y_i$ mark the cell position the robot moves to during step $i$. In A* search, potential path nodes are prioritized for exploration using an estimate of how long the resulting path would be. When a path to a goal is planned, the nodes are recorded in a shared ledger. The algorithm is ``cooperative" because agents use this shared information to avoid planning paths that:
\begin{enumerate}
    \item Include occupation of a cell at the same time another agent has already planned to occupy it,
    \item Include moving forward while the sensing cone is going to be occupied by another agent who planned earlier. The sensing cone always points in the direction of the next planned step,
    \item Include occupying a neighbor's sensing cone while the neighbor moves forward. 
\end{enumerate}

The resulting globally planned paths still respect the sensing cone rules. Agents do not move forward when their sensing cones are occupied, but due to advance planning instead of reaction to local stimuli. When agents reach their goals, they receive a new goal and plan a path to it. If needed, agents can wait in place. If an agent $a_i$ is planning but finds it can neither move anywhere nor wait in place (because another agent $a_j$ is already scheduled to move into $a_i$'s current position), $a_i$ will wait in place and force $a_j$ to replan.

Adjacent positions in a plan can be horizontal, vertical, or diagonal neighbors. To keep values rational so that timesteps can be discretized, a diagonal step in the planner takes 1.5x as long as a vertical or horizontal step. (The true length of a diagonal step is $\sqrt2 \approx 1.414$x longer than a horizontal step, so this introduces a $<10\%$ error.)

The two primary differences between our implementation and the original Cooperative A* algorithm presented in $\cite{silver2005cooperative}$ are the replanning (not needed in \cite{silver2005cooperative} because new goals are not assigned) and the incorporation of diagonal moves. 

We choose a cell width for A* which is small enough that if all agents are positioned at cell centers, an agent with sensing range $r$ can detect any neighbors in adjacent and diagonally adjacent cells. In the Fig.~\ref{fig:fig_intelligence} simulations, we use $L = 40$, $r = 2$, and a cell width of $1.33$ (so the arena is broken into $30 \times 30$ cells). We match the speeds of agents in the continuous setting and the Cooperative A* setting by determining that an agent moving horizontally for the same amount of time in both settings should travel the same distance $d$.

To verify that the global and local settings used in Fig.~\ref{fig:fig_intelligence} are comparable, we check the average goal attainment rates for single robots (so there are no effects from traffic) moving with no noise. In this case, goal attainment rate should match across the different motion controllers. The theory from Equation~\ref{eq:goal_travel_time} predicts a goal attainment rate of 0.0327. The observed attainment rate in simulation is 0.0296 goals per second for the global planner (9.4\% error from predicted value) and 0.034 goals per second for the local methods (4\% error). The error for the global planner comes from the 10\% error in how diagonal moves are handled. Error for the local methods occurs because robots reach the goal when they arrive within $\epsilon$ of the goal point, so they travel slightly less than the true goal distance. 

\section{Simulation Parameters}
\begin{table}[H]
\centering
\begin{tabular}{|l | l | l | p{1.9cm} | p{3.1cm} | p{1.9cm} | p{2.5cm} |} 
\hline
\textbf{Description}           & \textbf{Unit} & \textbf{Symbol} & \textbf{Noisy Local (Fig.~\ref{fig:fig2_theory})} & \textbf{Experiment Match (Fig.~\ref{fig:fig3_experiments})} & \textbf{Noisy Local (Fig.~\ref{fig:fig_intelligence})} & \textbf{Global Planner (Fig.~\ref{fig:fig_intelligence})} \\ \hline
Timestep Length &              s & $dt$                            & 0.1                                & 0.1                                                & 0.1                                & 2.667                      \\
Cruising Speed      & m/s           & $v$                             & 0.5                                & 0.1 (mean)                                            & 0.5                                & 0.5                              \\
Arena Sidelength    & m             & $L$                             & 40                                 & 1.2                                                & 40                                 & 40                               \\
Sensing Radius      & m             & $r$                             & 2                                  & 0.1564                                             & 2                                  & 2                                \\
Sensing Angle       & rad           & $\gamma$                        & $\frac{2\pi}{3}$                   & $\frac{2\pi}{3}$                                   & $\frac{2\pi}{3}$                   & $\frac{2\pi}{3}$                 \\
Avg. Step Length         & m             & $b$                             & 0.5                                & 0.15 (mean)                                             & 2.5                                & 1.333                             \\
Turn Speed       & rad/s         &                                 & $\infty$                           & 1.039 (mean)                                              & $\infty$                           & $\infty$                         \\
Goal Tolerance     & m             & $\epsilon$                      & 0.6                                & 0.08                                               & 0.6                                &                          \\
Trial Time   & s             &                                 & 8000                               & 300                                                & 8000                               & 8000                             \\
Boundaries &               &                                 & periodic                           & none                                               & periodic                           & periodic                         \\
Num. Trials & & & 20 or 50& 100 & 20 & 20 \\
\hline
\end{tabular}
\caption{\textbf{Parameters used for main text simulations.} For the noisy local simulations, simulations are run for 50 trials in the high-variance regime ($n \leq 128$ and $\sigma \leq 2.0$), and for 20 trials everywhere else. For the experiment match simulations, cruising speed and turn speed are matched to the average robot speed observed for the corresponding $(n, \sigma)$, which in turn also affects the step length. The values given here are averages across all $(n, \sigma)$ pairs.}
\label{table:parameters}
\end{table}

\end{document}